\documentclass[a4paper,fleqn]{cas-dc}
\usepackage[utf8]{inputenc}
\usepackage[numbers]{natbib}
\usepackage{multirow} 
\usepackage[table,xcdraw]{xcolor} 
\usepackage{makecell}
\usepackage{booktabs}
\usepackage{indentfirst}
\usepackage{caption}
\PassOptionsToPackage{a4paper,margin=2.5cm}{geometry} 
\usepackage{array}
\usepackage{graphicx}
\usepackage{boldline}
\usepackage{float}

\usepackage{bm} 
\usepackage{arydshln}
\usepackage{pifont}  
\usepackage{threeparttable}
\usepackage[font=normal,labelfont=bf]{caption}
\usepackage[all]{hypcap}
\newcommand{\cmark}{\textcolor{teal}{\ding{51}}}   
\newcommand{\xmark}{\textcolor{red}{\ding{55}}}    
\usepackage{tikz}

\newcommand*\circletextgray[1]{%
  \tikz[baseline=(char.base)]{
    \node[shape=circle,draw=gray,dashed,fill=gray!25,
          minimum size=12pt, inner sep=2pt,
          text height=1.5ex, text depth=.25ex] (char) {\small #1};
  }%
}

\newcommand*\circletext[1]{%
  \tikz[baseline=(char.base)]{
    \node[shape=circle,draw=gray,dashed,fill=white,
          minimum size=12pt, inner sep=2pt,
          text height=1.5ex, text depth=.25ex] (char) {\small #1};
  }%
}

\definecolor{lightyellow}{RGB}{255,255,224}
\definecolor{lightred}{RGB}{255,182,193}
\definecolor{lightpurple}{RGB}{216,191,216}
\definecolor{darkblue}{RGB}{123,166,180}
\definecolor{lightblue}{rgb}{0.92, 0.95, 1.0}
\definecolor{lightpink}{rgb}{1.0, 0.95, 0.95}
\usepackage{url}
\usepackage{adjustbox}
\PassOptionsToPackage{colorlinks,linkcolor=blue,anchorcolor=blue,citecolor=blue}{hyperref}

\def\tsc#1{\csdef{#1}{\textsc{\lowercase{#1}}\xspace}}
\tsc{WGM}
\tsc{QE}
\tsc{EP}
\tsc{PMS}
\tsc{BEC}
\tsc{DE}
\begin{document}
\setlength{\parindent}{2em}
\let\WriteBookmarks\relax

\def\floatpagepagefraction{1}
\def\textpagefraction{.001}

\shorttitle{}
\shortauthors{C. Wu et~al.}

\title [mode = title]{AnyAD: Unified Any-Modality Anomaly Detection in Incomplete Multi-Sequence MRI}




\author[1]{Changwei Wu}
\fnmark[1]

\affiliation[1]{organization={School of Computer Science and Technology, Hangzhou Dianzi University},
                city={Hangzhou},
                postcode={310018},
                state={Zhejiang},
                country={China}}

\author[2]{Yifei Chen}[style=chinese]
\fnmark[1]

\affiliation[2]{organization={School of Biomedical Engineering, Tsinghua University},
                postcode={100084},
                state={Beijing},
                country={China}}
                
\author[1]{Yuxin Du}

\author[2]{Mingxuan Liu}

\author[1]{Jinying Zong}

\author[3]{Beining Wu}

\affiliation[3]{organization={HDU-ITMO Joint Institute, Hangzhou Dianzi University},
                city={Hangzhou},
                postcode={310018},
                state={Zhejiang},
                country={China}}

\author[1]{Jie Dong}

\author[1]{Feiwei Qin}
\cormark[1]

\author[4]{Yunkang Cao}

\affiliation[4]{organization={School of Artificial Intelligence and Robotics, Hunan University},
                city={Changsha},
                postcode={410082},
                state={Hunan},
                country={China}}

\author[2]{Qiyuan Tian}
\cormark[1]

\fntext[fn1]{These authors contributed equally to this work.}

\cortext[cor1]{Corresponding authors: Feiwei Qin (qinfeiwei@hdu.edu.cn), Qiyuan Tian (qiyuantian@tsinghua.edu.cn).}



\begin{abstract}
Reliable anomaly detection in brain MRI remains challenging due to the scarcity of annotated abnormal cases and the frequent absence of key imaging modalities in real clinical workflows. Existing single-class or multi-class anomaly detection (AD) models typically rely on fixed modality configurations, require repetitive training, or fail to generalize to unseen modality combinations, limiting their clinical scalability. In this work, we present a unified Any-Modality AD framework that performs robust anomaly detection and localization under arbitrary MRI modality availability. The framework integrates a dual-pathway DINOv2 encoder with a feature distribution alignment mechanism that statistically aligns incomplete-modality features with full-modality representations, enabling stable inference even with severe modality dropout. To further enhance semantic consistency, we introduce an Intrinsic Normal Prototypes (INPs) extractor and an INP-guided decoder that reconstruct only normal anatomical patterns while naturally amplifying abnormal deviations. Through randomized modality masking and indirect feature completion during training, the model learns to adapt to all modality configurations without re-training. Extensive experiments on BraTS2018, MU-Glioma-Post, and Pretreat-MetsToBrain-Masks demonstrate that our approach consistently surpasses state-of-the-art industrial and medical AD baselines across 7 modality combinations, achieving superior generalization. This study establishes a scalable paradigm for multimodal medical AD under real-world, imperfect modality conditions. Our source code is available at \href{https://github.com/wuchangw/AnyAD}{https://github.com/wuchangw/AnyAD}.
\end{abstract}



\maketitle





\section{Introduction}
\label{sec:Introduction}
The field of medical diagnostics is undergoing a profound transformation, driven by the exponential growth of data, significant advances in artificial intelligence, and deep learning ~\citep{panahi2025deep}. Traditionally, clinicians have relied on clinical experience and conventional diagnostic tools. However, the massive scale and complexity of modern medical data have created a demand for sophisticated analytical methods, and deep learning has emerged as a promising solution \citep{guo2023encoder,huang2024adapting} due to its ability to learn intricate patterns and representations directly from raw data.
Despite the abundance of data, fully leveraging it remains challenging ~\citep{zhu2025xlstm,zhu2025no}. Anomalous samples still require manual identification and annotation, and such samples are often limited in number and variable in quality. Furthermore, due to data privacy concerns, obtaining well-annotated abnormal cases is even more difficult \citep{jeong2023winclip}.In the context of rare diseases characterized by highly diverse anomalies and extremely scarce samples, traditional supervised deep learning approaches face data limitations, as acquiring sufficient high-quality training data is often impractical, thereby constraining the development, generalization, and practical applicability of these techniques.

Unsupervised image anomaly detection (AD) ~\citep{xie2024iad,zhang2025comprehensive} aims to identify abnormal patterns in images by learning exclusively from standard samples and to localize anomalous regions further ~\citep{cao2024survey}. This approach relies solely on standard samples, which are relatively easy to obtain and abundant, eliminating the need for abnormal annotations and making it particularly suitable in scenarios where anomalous samples are scarce. Currently, AD techniques have been widely applied in industrial defect detection \citep{zou2022spot,bergmann2021mvtec} and video surveillance ~\cite{mabrouk2018abnormal}. Notably, methods such as Dinomaly, proposed by \cite{guo2025dinomaly}, and INP-Former, proposed by \cite{luo2025inp}, have demonstrated remarkable detection performance on industrial datasets, including MVTec AD and VisA.
However, significant differences exist between magnetic resonance imaging (MRI) data used in medical diagnostics and those in industrial datasets. Single-sequence MRI often cannot fully capture the detailed tissue structures and lesion characteristics, necessitating the use of multiple sequences for accurate diagnosis \citep{wang2025smart}. Common MRI modalities include T1, T2, and FLAIR. T1 reflects the brain's anatomical structure, and T2 is sensitive to water content. FLAIR, a specialized T2 sequence, is frequently used to diagnose brain edema and tumors \citep{liu2023deep}. It suppresses cerebrospinal fluid signals while retaining high-intensity signals in lesions, facilitating the detection of subtle abnormalities \citep{liu2023m3ae}.
In practical medical diagnostic scenarios, comprehensive assessment typically requires integrating information from multiple MRI modalities. However, factors such as regional limitations in healthcare infrastructure and patient financial constraints often result in missing sequences, posing additional challenges for automated multimodal analysis.

In traditional AD tasks, detection accuracy is often improved by adopting single-class AD approaches. Single-class models construct a separate model for each category, requiring retraining for every new class and producing non-transferable weight files. These weight files substantially increase storage demands, making deployment challenging in remote or resource-limited healthcare settings. Moreover, patients frequently have unpredictable missing MRI modalities. For new modality combinations, single-class models must be retrained to fully leverage the available information, further exacerbating storage and computational costs.

In recent years, to address the storage inefficiency associated with expanding the number of classes in single-class AD, traditional single-class \citep{roth2022towards} approaches have been extended to few-shot AD ~\citep{jiang2024prototypical} and multi-class AD \citep{luo2025inp,guo2025dinomaly}. Multi-class AD models employ a unified architecture to detect multiple classes simultaneously, requiring only a single set of generalizable weights, thereby significantly reducing storage demands and enabling the training and storage of numerous modality combinations. However, these methods still lack sufficient flexibility when faced with randomly missing MRI modalities and cannot handle arbitrary numbers of input modalities \citep{,zhu2025bridging}. Consequently, there is an urgent need for a universal architecture capable of accommodating any number of input modalities while balancing training complexity and storage efficiency.

\begin{table}[h]
\centering
\caption{\normalfont\normalsize C\textbf{omparison of different AD paradigms}. The table shows the characteristics of one-class, multi-class, and any-class AD. Our Any-class AD method can handle inputs of any modality and obtain universal training weights.}
\vspace{0.5em}
\label{table:paradigms_table}
\resizebox{\columnwidth}{!}{%
\begin{threeparttable}
\begin{tabular}{@{\hskip 0pt}l@{\hskip 6pt}ccc} 
\toprule
\textbf{Capabilities} & \textbf{Single-Class AD} & \textbf{Multi-Class AD} & \textbf{Any-Class AD} \\ 
\midrule
Single-modal Input          & \cellcolor{lightyellow!25} \cmark & \xmark & \cmark \\
Multi-modal Input          & \xmark & \cellcolor{lightyellow!25} \cmark & \cmark \\
Any-modal Input          & \xmark & \xmark & \cellcolor{lightyellow!25} \cmark \\
Weight Generalization        & \xmark & \xmark & \cellcolor{lightyellow!25} \cmark \\
\bottomrule
\end{tabular}
\end{threeparttable}
}
\end{table}

To address the challenges above, we propose an AD framework capable of handling any class and arbitrary modality combinations. This approach leverages a single, unified architecture and shared weights to accommodate inputs with varying numbers and types of modalities, significantly enhancing model generalization and input robustness. Unlike previous methods that are tailored to fixed modalities or specific classes, our framework maintains stable AD and precise localization even when presented with unseen modality combinations. This study not only extends the applicability of AD models to clinical multi-modal scenarios but also provides a novel paradigm for intelligent diagnostic systems under arbitrary input conditions.

The main contributions of this study are as follows:

• We present, for the first time, an AD framework capable of processing complex input combinations of an arbitrary number and modality through a single unified architecture, shared weights, and a one-time training procedure.

• We introduce an innovative indirect feature completion mechanism that enhances weak modality combinations by guiding the reconstruction of missing modalities at the feature level, thereby improving detection accuracy for suboptimal modality configurations.

• We incorporate a prototype-based learning mechanism to enhance model generalization, embedding prototypes into the feature reconstruction process to mitigate false anomaly predictions caused by shortcut learning.

\section{Related Work}
\subsection{Research on Single-Class Anomaly Detection}
\label{sec:related_work:single_class}
Traditional AD methods are based mainly on single-class AD, initially proposed by MVTec-AD \citep{bergmann2021mvtec}. The core idea is to build a separate anomaly-detection model for each class. Techniques such as reconstruction-based ~\citep{luo2024ami,zhang2024realnet}, prototype-based \citep{roth2022towards}, and embedding-based \citep{liu2023simplenet} methods enable the model to learn the characteristics of normal images. \cite{zhao2023ae} proposed AE-FLOW, an autoencoder with a normalized flow bottleneck, which constructs both a loss function and an anomaly scoring function by combining the advantages of normalized flow methods for estimating image-level feature anomaly likelihoods with the interpretability of pixel-level reconstruction-based approaches. ~\cite{zhang2024anomaly} employed a dynamic gating strategy to manage skip connections in reconstruction-based AD methods. They introduced a novel gated highway connection module to adaptively integrate skip connections into the framework, thereby improving AD performance. While these models achieve excellent results in single-class AD tasks, practical medical diagnostic scenarios present additional challenges. As shown in Table~\ref{table:paradigms_table}, the limited representational capacity of single modalities and the absence of specific modalities \citep{liu2023m3ae} necessitate integrating multiple modalities for accurate diagnosis. Due to missing modalities, numerous modality combinations arise, requiring models to be trained separately for each combination \citep{luo2025exploring}, with weights that cannot be reused. This significantly increases storage requirements, posing challenges for deployment in resource-limited or grassroots healthcare settings.

\subsection{Research on Multi-Class Anomaly Detection}
\label{sec:related_work:multi_class}
To address the high resource consumption of single-class AD, researchers have proposed multi-class AD frameworks \citep{you2022unified}. Multi-class AD employs a unified model with shared weights to detect multiple classes, effectively reducing resource demands. \cite{lu2023hierarchical} proposed HVQ-Trans, which leverages a vectorized framework to mitigate shortcut learning issues. \cite{qin2025mambaad} introduced the MambaAD model, an efficient anomaly-detection framework based on the linear-time state-space model Mamba. ~\cite{guo2025dinomaly} presented Dinomaly, which uses the medical foundation model DINO \citep{darcet2023vision} to extract shallow image features and subsequently reconstructs them. By employing a scalable base Transformer architecture along with simple operations such as Dropout, Dinomaly achieves flexible reconstruction without enforcing strict layer-wise or point-wise constraints. Even with a relatively simple architecture, Dinomaly attains high detection performance, substantially enhancing the capabilities of multi-class AD models. However, existing multi-class AD methods remain unable to accommodate arbitrary input classes. In brain tumor detection tasks, patients often present with missing MRI modalities, resulting in highly variable numbers and types of input modalities. Traditional multi-class methods often lack the flexibility to handle such complex scenarios and may require modality imputation or even retraining, thereby limiting their applicability in real-world clinical settings.

\subsection{Research on Prototype Learning}
\label{sec:related_work:prototype_learning}
Prototype learning \citep{snell2017prototypical} aims to extract representative prototypes from a given training set and use them to measure sample distances in a metric space for classification. This technique has been widely applied in few-shot learning \citep{li2021adaptive}. \cite{roth2022towards} proposed PatchCore, which extracts multiple standard prototype combinations to represent the normality of the training data. During testing, it maximizes the use of available raw information while minimizing bias toward specific categories in the dataset. \cite{park2020learning} introduced MNAD, incorporating prototypes into the reconstruction task to mitigate shortcut learning and enhance model robustness. \cite{luo2025exploring} proposed INP-Former, an advanced prototype-based model that effectively presents the typical vector prototype extractor and further integrates intrinsic normal prototypes (INPs) into a reconstruction-based AD framework. This innovative design endows INP-Former with strong generalization capabilities, allowing it to robustly extract INPs even from previously unseen classes. The exceptional generalization capability of INP-Former provides a highly promising and practical direction for effectively addressing challenging AD tasks in realistic medical scenarios with partially missing modalities.

To address the limitations of traditional multi-class AD methods in handling arbitrary modality inputs, we propose an AD framework capable of processing inputs of any class and modality combination. Building upon the strengths of the INP-Former framework, we develop a high-input-robustness model tailored for brain tumor detection across arbitrary class-modality scenarios.
To accommodate the complexity of arbitrary modality inputs, we introduce a novel training paradigm and model enhancement strategy. During training, modality absence is actively simulated, compelling the model to learn to adapt to any combination of input modalities. This enables the model to generalize to scenarios with extensive missing modalities, including previously unseen modality combinations. Additionally, we propose an indirect feature-completion strategy that guides the model during training to reconstruct missing modality features. As a result, during testing, the model automatically completes missing modality features. For example, using low-quality single-modal T1 input can achieve detection performance comparable to that of high-quality modalities. By leveraging information from multiple modalities, this approach enhances detection performance for suboptimal modality combinations, resulting in a model that adapts to arbitrary class and modality inputs without requiring separate training for each missing modality scenario.

\section{Method}

\begin{figure*}
\centering
\includegraphics[width=\linewidth]{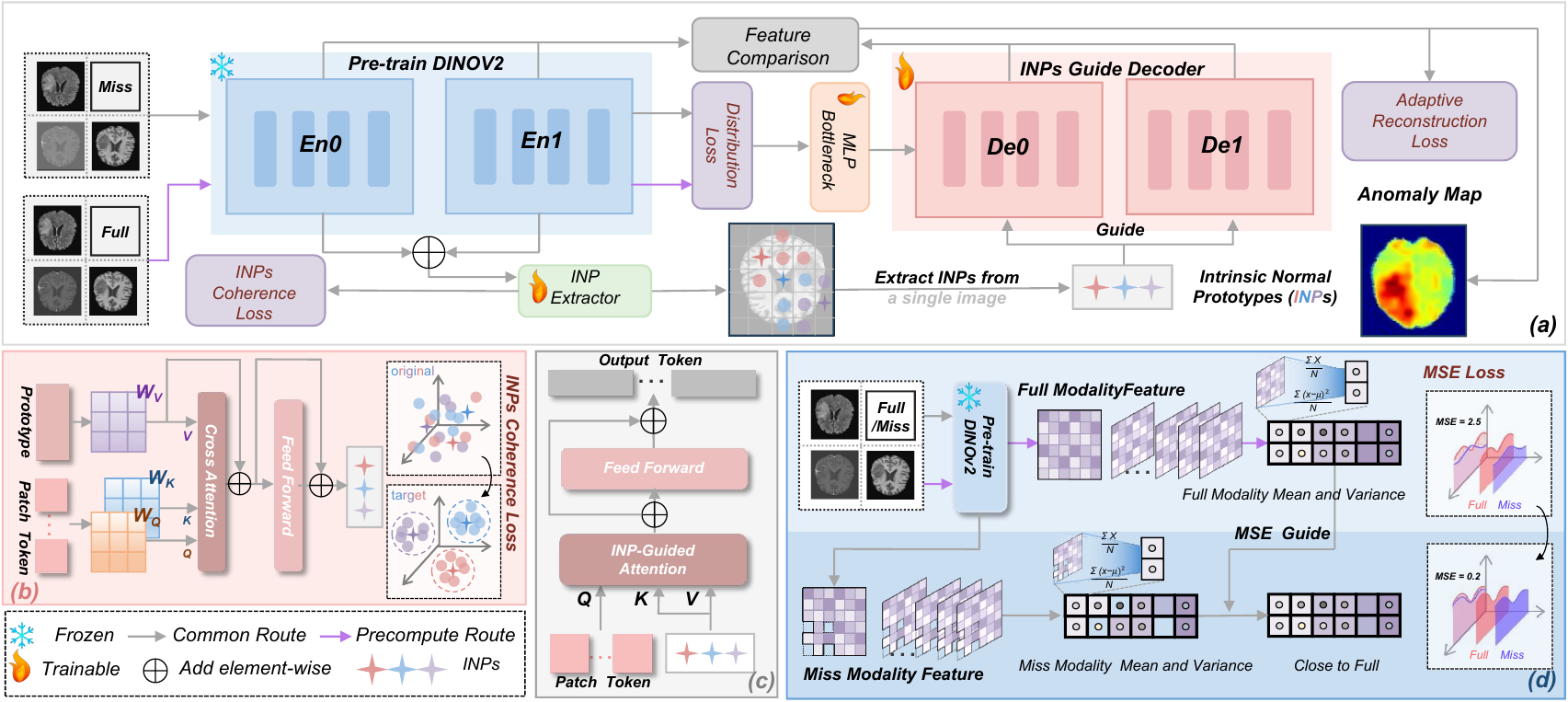}  
\caption{\normalfont\normalsize\textbf{Overview of our AnyAD framework.} (a) Our model consists of a pre-trained encoder, an INPs extractor, a bottleneck, and an INPs-guided decoder. Simultaneously, the dual paths extract the means and variances of the full-modal features and the missing-modal features, respectively. (b) Detailed process of INPs extractor and INPs consistency loss. (c) Detailed architecture of the INPs-guided decoder. (d)Detailed process of feature distribution alignment loss. }
\label{fig:model_arch}
\end{figure*}

\subsection{Overview}
\label{sec:method:Overview}
In this study, we propose an Any-Modality feature distribution alignment framework to achieve robust AD across arbitrary modality combinations. As shown in Fig. \ref{fig:model_arch}, the framework integrates a dual-pathway pre-trained encoder with a distribution alignment strategy. The encoder extracts multi-scale features and generates two sets of fused feature maps, which are used for anomaly contrast and reconstruction, respectively. The distribution alignment strategy minimizes statistical differences between missing-modality and full-modality features, ensuring consistent feature distributions and detection performance even under incomplete modality inputs. Additionally, we incorporate an INPs extractor and an INPs-guided decoder that reconstruct features using only normal prototypes, thereby explicitly exposing anomalies through discrepancies between the encoded and decoded features. During training, partial modalities are randomly masked to simulate realistic missing-modality scenarios, and the model is jointly optimized with multiple loss functions. This design ensures stable generalization and robust AD across any modality combination.

\subsection{Any-Modality Feature Distribution Alignment}
\label{sec:method:Any-Modality Feature Distribution Alignment}
The core of the proposed model is the Any-Modality Feature Distribution Alignment Framework, which enables robust AD across arbitrary modalities. The architecture comprises a dual-pathway pre-trained encoder and a distribution alignment strategy. The encoder efficiently extracts and fuses multi-scale features. At the same time, the distribution alignment mechanism explicitly constrains the feature space, ensuring that the distribution of features for missing modalities approximates that of the full-modality case. This significantly enhances the model’s generalization and robustness under incomplete multi-modal inputs.


\subsubsection{Dual-pathway pre-trained encoder}
\label{sec:method:Dual-pathway pre-trained encoder}
The dual-pathway pre-trained encoder serves as the foundational module of the framework. We employ a Transformer with DINOv2 ViT-Base pre-trained weights as the feature extractor to capture semantic information across multiple image scales fully. After passing the input through the encoder, multi-layer feature maps are obtained, which are then divided into two groups $\boldsymbol{F}_{\boldsymbol{deep}}$ and $\boldsymbol{F}_{\boldsymbol{shallow}}$ based on feature hierarchy. Each group is fused separately to produce two fused feature maps, $\boldsymbol{En0}$ and $\boldsymbol{En1}$. For an input image $\boldsymbol{I} \in \mathbb{R}^{B \times C \times H \times W}$, this process can be formalized as follows:
\begin{flalign}
& \begin{aligned}
&\bm{\mathcal{F}}_{\ell} = E(I), \\
&En_0 = \bm{\mathcal{F}}\mathrm{usion}(\bm{\mathcal{F}}_{\text{shallow}}), 
En_1 = \bm{\mathcal{F}}\mathrm{usion}(\bm{\mathcal{F}}_{\text{deep}}),
\end{aligned} &
\label{eq:fusion}
\end{flalign}
where, $\bm{\mathcal{F}}_{\ell}$ denotes the feature map from the $l$-th layer, $E(\cdot)$ represents the pre-trained encoder, and $\bm{\mathcal{F}}\mathrm{usion}(\cdot)$ denotes the feature fusion operation. The two fused feature maps, $En_0$ and $En_1$, serve distinct functional pathways. In the AD pathway, $En_0$ and $En_1$ are cached for subsequent comparison with decoder-generated features, enabling the identification of anomalous regions. In the decoder input pathway, $En_0$ and $En_1$ are further processed through a bottleneck layer for high-dimensional feature fusion. The bottleneck first performs element-wise addition of the two feature maps, expands the channel dimension from 768 to 3072, and then compresses it back to 768 channels. This process can be formalized as follows:
\begin{flalign}
& \begin{aligned}
&\bm{\mathcal{F}}_{\text{high}} = \mathrm{Linear}_{768\to3072}\bigl( En_0 \oplus En_1 \bigr) , \\
&\bm{\mathcal{F}}_{\text{bottleneck}} = \mathrm{Linear}_{3072\to768}\bigl( \bm{\mathcal{F}}_{\text{high}} \bigr) ,
\end{aligned} &
\label{eq:bottleneck}
\end{flalign}
where, $\mathrm{Linear}_{in\to out}$ denotes a fully connected layer that performs independent channel-wise transformation, and $\oplus$ represents element-wise addition. This up-projection, followed by a subsequent compression operation, enhances feature separability in the high-dimensional space, thereby further improving the representational capacity of the encoded features and the overall quality of downstream reconstruction.

\subsubsection{Distribution alignment strategy}
To ensure robust performance across arbitrary modalities, partial modalities are randomly masked during training to simulate missing-modality scenarios, thereby enhancing the model’s adaptability to incomplete inputs. However, relying solely on random masking does not guarantee distribution consistency. To address this, we introduce a distribution alignment loss on top of the dual-pathway encoder. Before training, we perform a pre-computation step on full-modality images: we use the fused feature maps from all modalities to compute the channel-wise mean and variance, which are cached as reference distributions. This step does not involve any parameter updates and therefore does not increase training overhead. During training, the input may contain missing modalities, and the model concurrently computes the statistical distribution of the current features. The distribution alignment loss then constrains the difference between the current and reference distributions to be minimal. This ensures that features from missing modalities are aligned to the full-modality feature space at the statistical level, significantly improving detection performance and feature stability. Specifically, for a given feature map, we compute the mean and variance along the channel dimension for each channel as follows:
\begin{equation}
\begin{aligned}
\mu_c &= \frac{1}{BHW} \sum_{b=1}^{B} \sum_{h=1}^{H} \sum_{w=1}^{W} \bm{\mathcal{F}}_{b,c,h,w}, \\
\sigma_c^2 &= \frac{1}{BHW} \sum_{b=1}^{B} \sum_{h=1}^{H} \sum_{w=1}^{W} 
\bigl(\bm{\mathcal{F}}_{b,c,h,w} - \mu_c\bigr)^2,
\end{aligned}
\label{eq:mean_var}
\end{equation}
where, $b$ denotes the batch index, $c$ the channel index, $h$ and $w$ represent the spatial height and width indices of the feature map. ${\mu_c}$ and ${\sigma_c}^2$ correspond to the mean and variance of each channel, respectively. The distribution alignment loss is then defined as follows (see equation ~\ref{eq:dist_loss}). This loss is optimized via backpropagation, allowing the model to consistently converge to a similar feature distribution space across different modality combinations, thereby maintaining stable AD performance. This design effectively mitigates feature shift caused by missing modalities and enhances the model's generalization and robustness.

\subsection{INP-Former}
To further ensure the quality and semantic consistency of decoded features, we retain the core principles of INP-Former within the overall architecture, including the INPs extractor and the INPs-guided reconstruction decoder. This design enables the model to reconstruct solely based on typical patterns, thereby explicitly revealing anomalous regions through reconstruction discrepancies.

\subsubsection{INPs extractor}
Prototype extraction is primarily achieved through an INPs extractor. Specifically, the process begins with randomly initialized prototypes, which are iteratively refined to obtain representative intrinsic normal prototypes (INPs). Concretely, the randomly initialized prototypes and the fused features from the pre-trained encoder are each projected using learnable weight parameters to obtain queries ($\boldsymbol{Q}$), keys ($\boldsymbol{K}$), and values ($\boldsymbol{V}$). Cross-attention is then applied to $\boldsymbol{Q}$, $\boldsymbol{K}$, and $\boldsymbol{V}$, and the resulting attention output is added to the initial random prototypes to form the attention prototypes. Finally, the attention prototypes are further updated through a feedforward neural network and added to the previous attention prototypes, producing the final representative prototypes that capture intrinsic regular features, called INPs:
\begin{equation}
\begin{aligned}[t]  
&\bm{\mathcal{Q}} = \bm{\mathcal{P}}_0 \, \bm{W}^{\mathcal{Q}}, \quad
\bm{\mathcal{K}} = \bm{\mathcal{F}}_Q \, \bm{W}^{\mathcal{K}}, \quad
\bm{\mathcal{V}} = \bm{\mathcal{F}}_Q \, \bm{W}^{\mathcal{V}}, \\[1mm]
&\bm{\mathcal{P}}' = \mathrm{Attention}(\bm{\mathcal{Q}}, \bm{\mathcal{K}}, \bm{\mathcal{V}}) + \bm{\mathcal{P}}_0, \quad
\bm{\mathcal{P}} = \mathrm{FFN}(\bm{\mathcal{P}}') + \bm{\mathcal{P}}',
\end{aligned}
\label{eq:p_attention}
\end{equation}
where, $\boldsymbol{\mathcal{W}^{\mathcal{Q}}}$, $\boldsymbol{\mathcal{W}^{\mathcal{K}}}$, and $\boldsymbol{\mathcal{W}^{\mathcal{Q}}}$ are the learnable weight parameters for $\boldsymbol{\mathcal{Q}}$, $\boldsymbol{\mathcal{K}}$, and $\boldsymbol{\mathcal{V}}$; $\boldsymbol{P}_0$ denotes the randomly initialized prototypes; and $\boldsymbol{\mathcal{F}}_Q$ represents the fused features from the pre-trained encoder. ${Attention}(\cdot)$ denotes the cross-attention operation, and $\mathrm{FFN}(\cdot)
$ represents the feedforward neural network. $\boldsymbol{\mathcal{P}}$ denotes the final prototypes. Additionally, a prototype consistency loss is employed (see equation ~\ref{eq:p_consistency_loss}) to ensure that the INPs consistently represent regular features throughout the extraction process, minimizing the risk of capturing general features and maintaining the specificity of the prototypes to standard patterns.

\subsubsection{INPs Guided Reconstruction}
The updated prototypes $P$ are used to guide the decoder in feature reconstruction. The decoder is composed of multiple INPs-guided decoding blocks. Within each decoding block, the upsampled features interact with $P'$. Specifically, the decoding features serve as queries ($\boldsymbol{Q}_L$), while the INPs act as keys ($\boldsymbol{K}_L$) and values ($\boldsymbol{V}_L$) through a specialized prototype attention mechanism. This design enables the decoder to reference the learned normal prototypes when reconstructing image features, allowing accurate reconstruction of normal tissue while producing poorer reconstruction for anomalous regions that do not conform to any prototype:
\begin{flalign}
& \begin{aligned}
&\bm{\mathcal{Q}}_{\ell} = \bm{\mathcal{F}}_D^{\,\ell-1} \, \bm{W}^{\mathcal{Q}_{\ell}}, \quad
\bm{\mathcal{K}}_{\ell} = \bm{\mathcal{P}} \, \bm{W}^{\mathcal{K}_{\ell}}, \quad
\bm{\mathcal{V}}_{\ell} = \bm{\mathcal{P}} \, \bm{W}^{\mathcal{V}_{\ell}}, \\[1mm]
&\bm{\mathcal{F}}_D^{\,\ell-1'} = \mathrm{ReLU}\!\left( \bm{\mathcal{Q}}_{\ell} \, \bm{\mathcal{K}}_{\ell}^\top \right) \bm{\mathcal{V}}_{\ell}, \quad
\bm{\mathcal{F}}_D^{\,\ell} = \mathrm{FFN}\!\left( \bm{\mathcal{F}}_D^{\,\ell-1'} \right) + \bm{\mathcal{F}}_D^{\,\ell-1'},
\end{aligned} &
\label{eq:d_attention}
\end{flalign}
where, 
$\boldsymbol{W}^{\boldsymbol{\mathcal{Q}}_{\ell}}$, 
$\boldsymbol{W}^{\boldsymbol{\mathcal{K}}_{\ell}}$, 
and 
$\boldsymbol{W}^{\boldsymbol{\mathcal{V}}_{\ell}}$ 
are the learnable weight parameters for 
$\bm{\mathcal{Q}}_{\ell}$, 
$\bm{\mathcal{K}}_{\ell}$, 
and 
$\bm{\mathcal{V}}_{\ell}$, 
and 
$\bm{\mathcal{F}}_D^{\,\ell}$ denotes the feature map from the $L$-th decoding layer. 
ReLU represents the ReLU activation function. 
Similarly, the decoded features are divided into $De_0$ and $De_1$, which are fused separately and subsequently used for feature comparison with the corresponding features extracted by the pre-trained encoder.

\subsection{Loss function}
\subsubsection{Prototype Consistency Loss}
The prototype consistency loss enforces that the INPs consistently represent the normal features of images by minimizing the cosine distance between each image feature and its nearest corresponding INP across the dataset. This ensures stable semantic referencing for the model and preserves the high-fidelity representation of normal features.
\begin{equation}
\begin{aligned}
&\bm{d}_i = \min_{n \in \{1, \dots, N\}} 
\bm{\mathcal{C}}\mathrm{osSim}\bigl(\bm{\mathcal{F}}_Q(i), \bm{\mathcal{P}}_n\bigr), \\[1mm]
&\bm{\mathcal{L}}_{\text{con}} = \frac{1}{N} \sum_{i=1}^{N} \bm{d}_i,
\end{aligned}
\label{eq:p_consistency_loss}
\end{equation}
where, $\bm{\mathcal{C}}\mathrm{osSim}(\cdot)$ denotes cosine similarity, $\boldsymbol{\mathcal{F}}_Q(i)$ represents the feature at the $i$-th position, $\boldsymbol{\mathcal{P}}_n$ denotes the $n$-th INP, $\boldsymbol{d}_i$ is the cosine distance between the feature and its nearest INP, and $N$ is the number of INPs.

\subsubsection{Adaptive Reconstruction Loss}
The adaptive reconstruction loss serves as the primary loss function. It computes the ratio of the minimum distance between the current sample feature and the INPs to the average cosine distance within the current batch, then applies a temperature hyperparameter to derive a difficulty weight for each sample. This mechanism allows the model to focus on regions that are difficult to reconstruct adaptively. The final loss is obtained by computing the cosine distance between the corresponding features of the pre-trained encoder and the INPs-guided decoder, multiplying each distance by its difficulty weight, summing over all features across channels and spatial locations, and averaging across the batch.
\begin{equation}
\begin{aligned}
&\omega_i = \left( \frac{\bar{d}}{d_i} \right)^{\gamma},\\[2mm]
&\bm{\mathcal{L}}_{\text{rec}} = \frac{1}{2} \sum_{i=0}^{L} 
\Bigl( \bm{\mathcal{C}}\mathrm{osSim}\bigl(\bm{\mathcal{F}}_{en_i}, \bm{\mathcal{F}}_{de_{i-1}}\bigr) \cdot \omega_i \Bigr),
\end{aligned}
\label{eq:adaptive_reconstruction_loss} 
\end{equation}
where ${\omega}_i$ denotes the adaptive weight for explicitly emphasizing difficult-to-reconstruct regions, $\bar{{d}}$ represents the average cosine distance, and $\gamma$ is the temperature hyperparameter, set to a default value of 3.

\subsubsection{Distribution Alignment Loss}
Before training, a pre-computation step is performed to calculate and store the mean and variance of the full-modality features. During actual training, the mean and variance of features from inputs with missing modalities are likewise computed. A distribution loss is then introduced to minimize the statistical discrepancy between the missing-modality features and the full-modality reference, thereby promoting alignment in the high-dimensional feature space.
\begin{equation}
\bm{\mathcal{L}}_{\text{dist}} 
= \mathrm{MSE}(\mu_{\text{current}}, \mu_{\text{full}})
+ \mathrm{MSE}(\sigma^2_{\text{current}}, \sigma^2_{\text{full}}),
\label{eq:dist_loss}
\end{equation}
where, ${\mu}_{{current}}$ and $\boldsymbol{\sigma}^2_{\boldsymbol{current}}$ denote the mean and variance of the current possibly incomplete modalities, while ${\mu}_{{full}}$ and $\boldsymbol{\sigma}^2_{\boldsymbol{full}}$ represent the mean and variance of the full-modality features. $\mathrm{MSE}(\cdot)$ denotes the mean squared error. 

The overall training loss of the model is a weighted combination of these individual losses, which is jointly optimized and backpropagated to update all 
\begin{equation}
\bm{\mathcal{L}}_{\text{total}} 
= \bm{\mathcal{L}}_{\text{rec}} 
+ \lambda_1 \cdot \bm{\mathcal{L}}_{\text{con}} 
+ \lambda_2 \cdot \bm{\mathcal{L}}_{\text{dist}},
\label{eq:total_loss}
\end{equation}
where, $\lambda_1$ is set to 0.2 by default, and $\lambda_2$ was determined through hyperparameter tuning experiments, with a value of 0.2 yielding the best model performance.

\begin{figure*}[t]
\centering
\includegraphics[width=\textwidth]{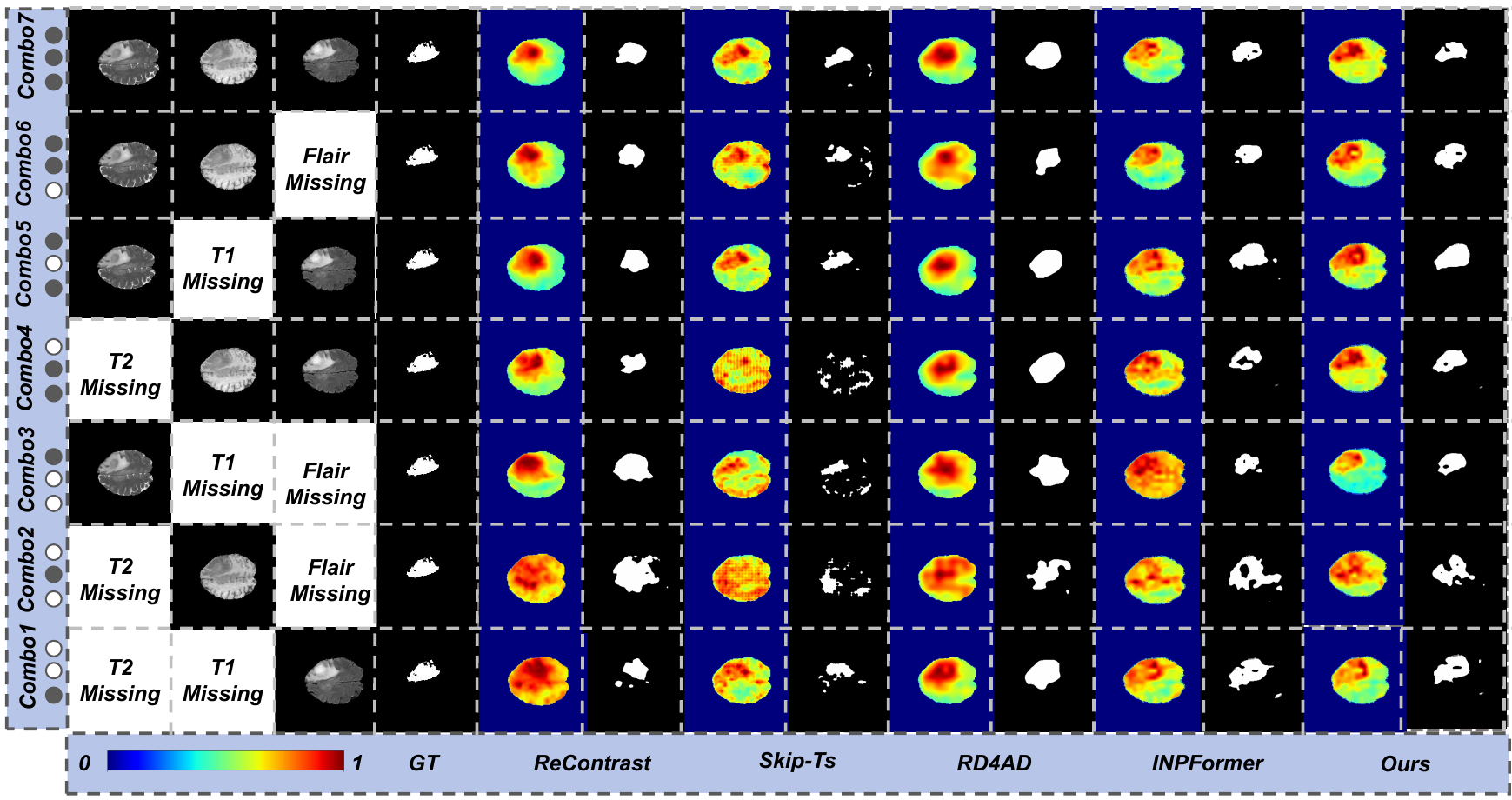}  
\caption{\normalfont\normalsize\textbf{Anomaly localization visualizations of the top-5 models on the BraTS2018 dataset.} Each row corresponds to one modality combination, numbered from 1 to 7, with detailed compositions listed in the first three columns. This convention for referring to modality combinations is consistent throughout the paper. From left to right, the columns show: the modality combination details, the ground truth segmentation mask as reference standard, and alternating rows of anomaly heatmaps and model-predicted localization maps, where odd rows display heatmaps and even rows show prediction maps.}
\label{fig:BraTS Heat Map}
\end{figure*}

\begin{figure*}[t]
\centering
\includegraphics[width=\textwidth]{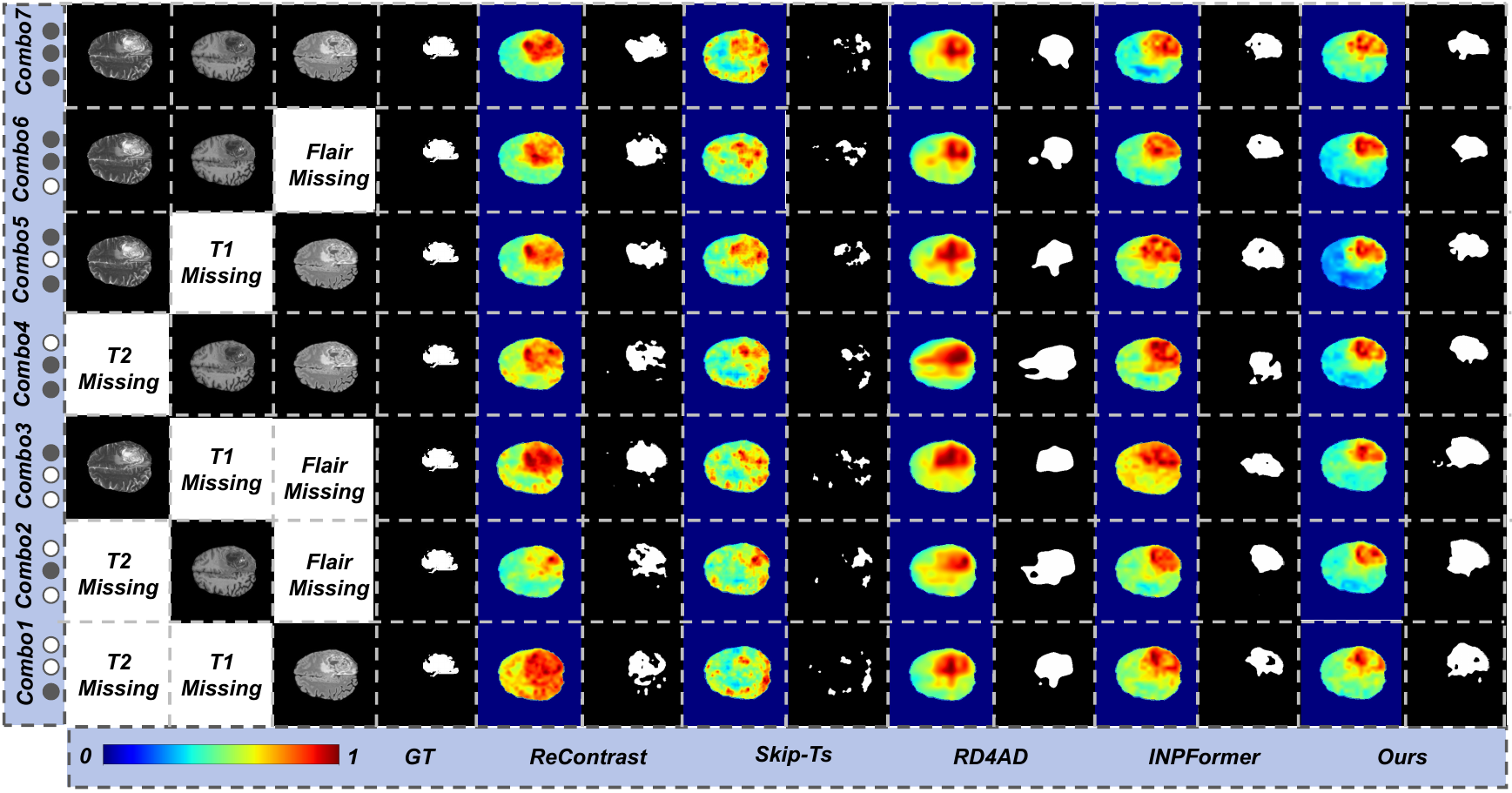}
\caption{\normalfont\normalsize\textbf{Anomaly localization visualizations of the top-5 models on the MU-Glioma-Post dataset.} Each row corresponds to one modality combination, numbered from 1 to 7, with detailed compositions listed in the first three columns. This convention for referring to modality combinations is consistent throughout the paper. From left to right, the columns show: the modality combination details, the ground truth segmentation mask as reference standard, and alternating rows of anomaly heatmaps and model-predicted localization maps, where odd rows display heatmaps and even rows show prediction maps.}
\label{fig:MU Heat Map}
\end{figure*}

\begin{figure*}[t]
\centering
\includegraphics[width=\textwidth]{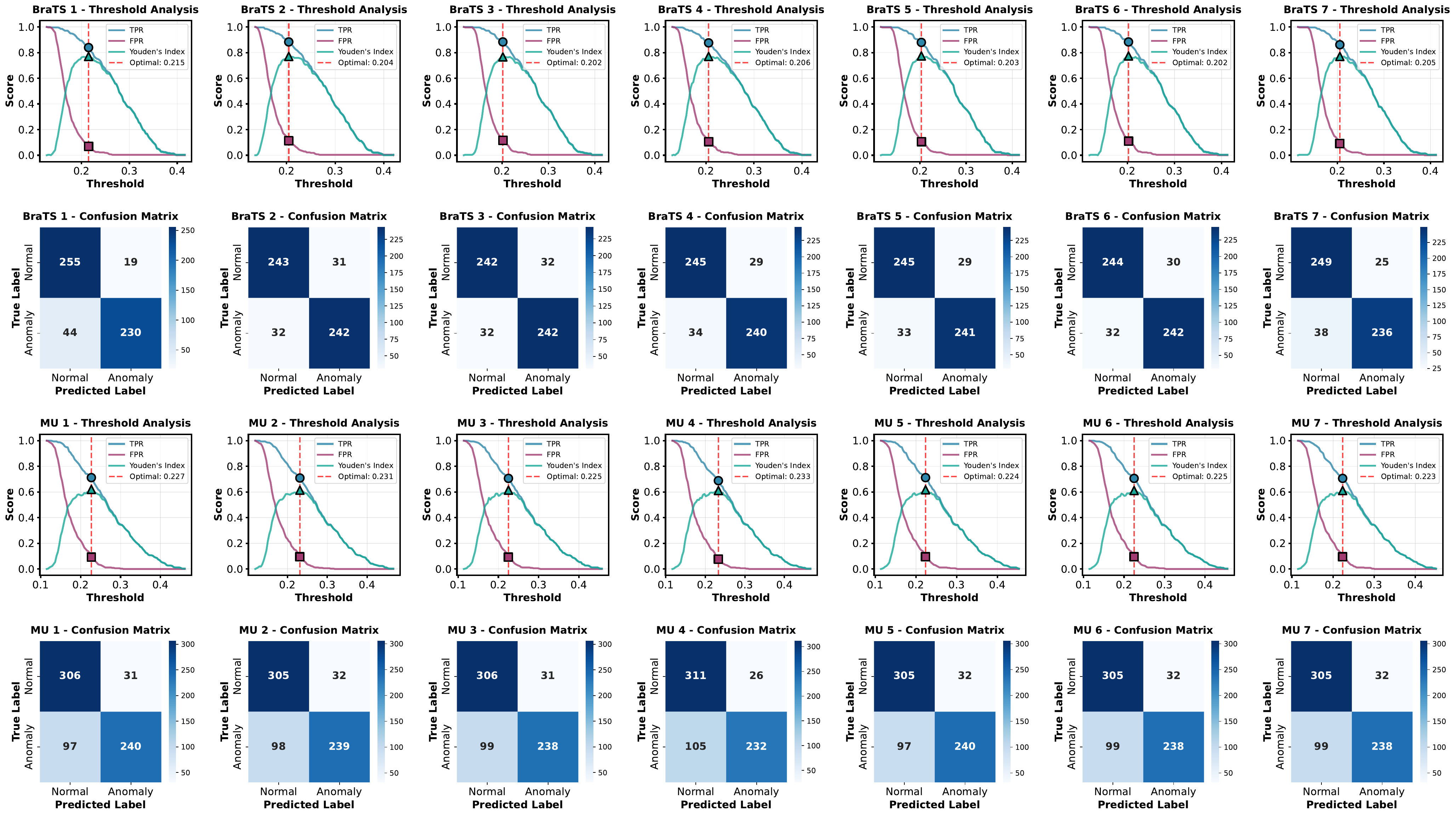}
\caption{\normalfont\normalsize
\textbf{Threshold curves and confusion matrices of Any-AD on BraTS2018 (top two rows) and MU-Glioma-Post (bottom two rows) datasets.} From left to right, the columns correspond to modality combinations 1-7.
}\label{fig:Confusion Matrix}
\end{figure*}

\section{Experiments}
\subsection{Dataset}

\begin{figure*}[]
\centering
\includegraphics[width=\textwidth]{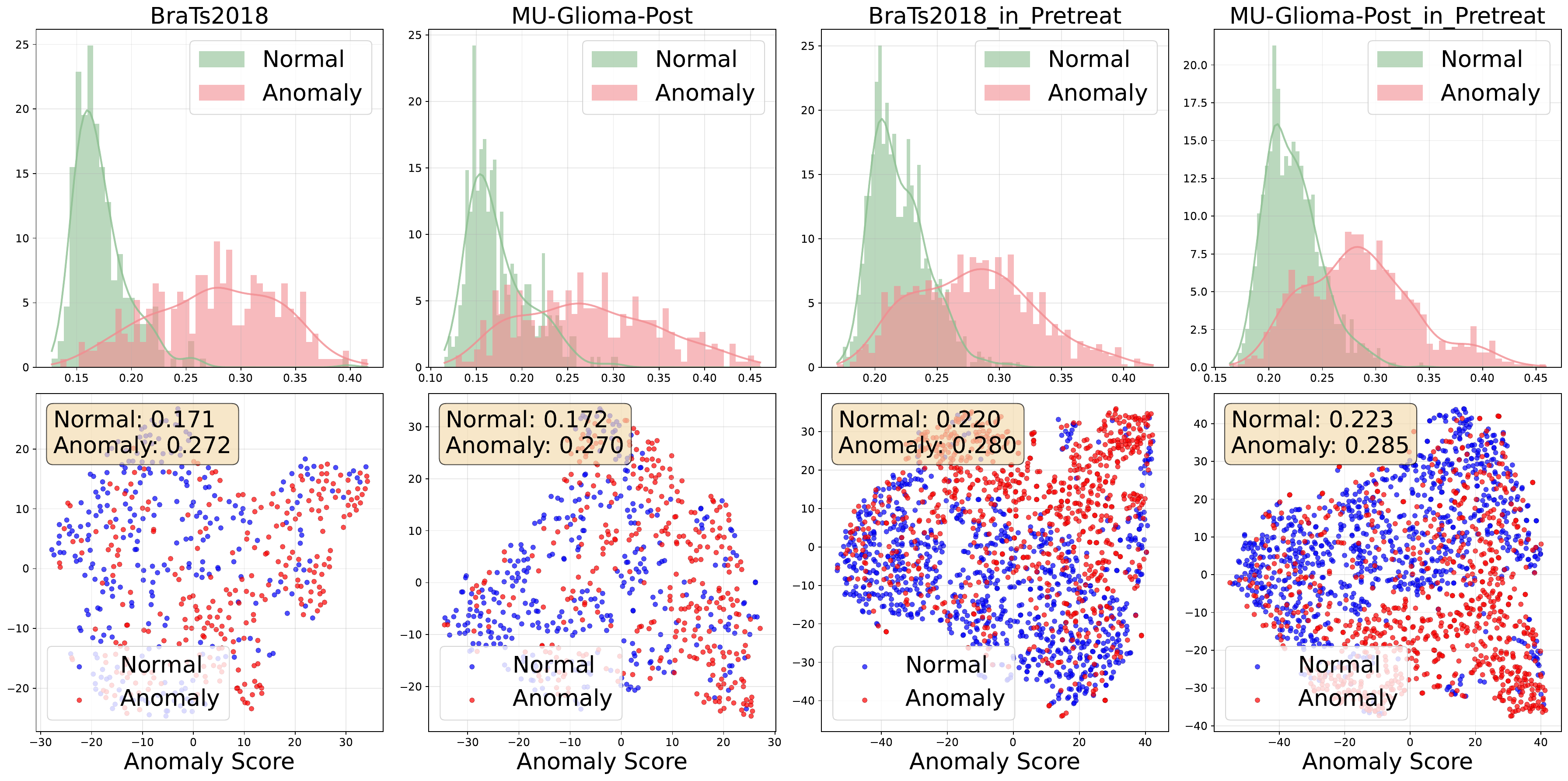}  
\caption{\normalfont\normalsize\textbf{Distribution and low-dimensional representation of anomaly scores}. Top row: histograms of anomaly scores; bottom row: t-SNE visualization of the scores.}                   
\label{fig:anomaly_score_distributions in different dataset}
\end{figure*}

\begin{figure*}[]
\centering
\includegraphics[width=\textwidth]{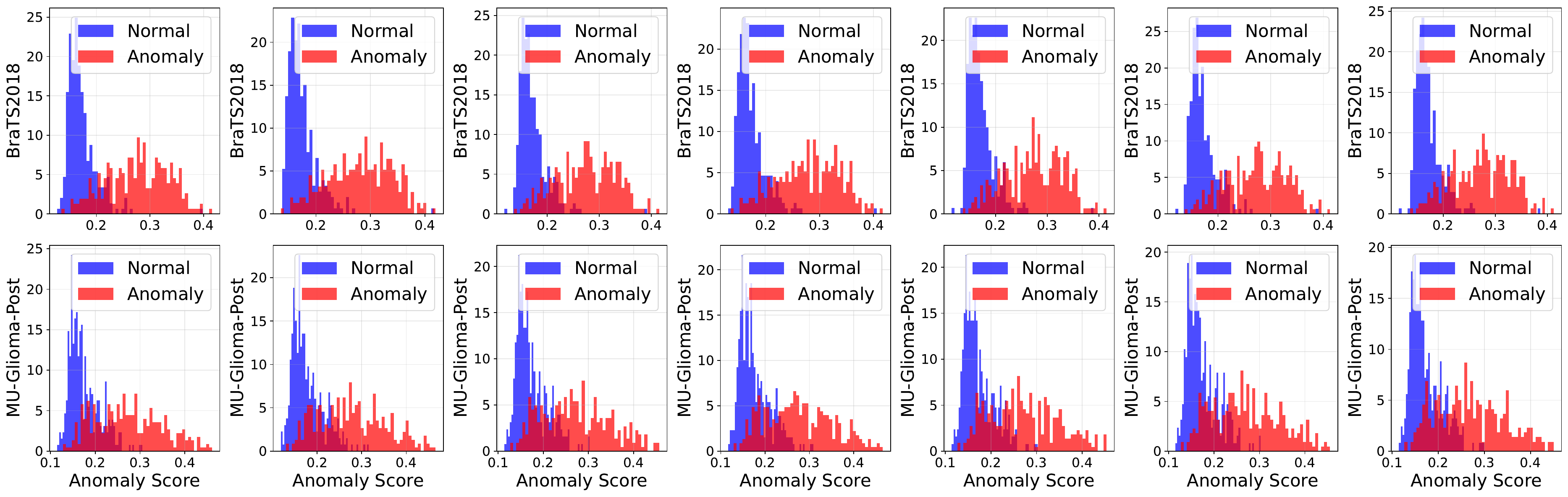}

\caption{\normalfont\normalsize\textbf{Anomaly score distributions of different modality combinations on BraTS2018 (top row) and MU-Glioma-Post (bottom row) datasets}. From left to right, the columns correspond to modality combinations 1-7.}                   
\label{fig:anomaly_score_distributions in different combanation}
\end{figure*}

\subsubsection{BraTS2018}
The BraTS2018 dataset\cite{menze2014multimodal} comprises 285 cases of high-grade gliomas (HGG) and 75 cases of low-grade gliomas (LGG), each including three-dimensional, multi-modal MRI scans accompanied by expert annotations. Each case contains four imaging modalities: T1, T2, T1 with contrast enhancement (T1Gd), and FLAIR. All lesion masks were meticulously delineated by neuroradiology experts, distinguishing the enhancing tumor region (ET), the tumor core (TC), and the whole tumor region (WT). In order to maintain sample consistency and focus on the AD task, this study only used 285 cases of HGG for experimental analysis.

\subsubsection{MU-Glioma-Post}
The MU-Glioma-Post dataset\cite{de20242024} originates from the University of Missouri Hospital and comprises 203 pathologically confirmed glioma patients, totaling 617 postoperative MRI time points. The imaging data were processed through a standardized preprocessing pipeline, and automatic tumor segmentation was performed using the nnUNet model, with the results subsequently reviewed and corrected by neuroradiology experts. The dataset maintains consistency with BraTS2018 in terms of modality configuration and label definitions; however, as it is derived from postoperative cases, the images contain complex structures such as surgical cavities, scarring, and radiotherapy-induced changes. Consequently, the segmentation outcomes of MU-Glioma-Post differ markedly from those of BraTS2018, clearly reflecting increased tumor heterogeneity and substantial variability in image characteristics, and thus presenting even greater challenges for accurate and robust detection.

\subsubsection{Pretreat-MetsToBrain-Masks}
The Pretreat-MetsToBrain-Masks dataset \citep{ramakrishnan2024large} aggregates high-resolution 3D MRI scans from 200 patients with brain metastases, sourced from the Yale New Haven Health System, the Yale Tumor Registry, and the Yale Gamma Knife Registry. All lesion regions were manually segmented by radiology experts and subsequently carefully standardized. The dataset retains modality configurations and labeling conventions consistent with BraTS2018; however, owing to its heterogeneous origins and the complexity of the lesions, which include sub-centimeter small lesions and necrotic tumor regions, it exhibits pronounced cross-domain variations in spatial distribution and imaging characteristics. These factors impose significantly elevated and stringent generalization requirements on current AD models.

\subsection{Data Preprocessing}
All imaging data employed in this study consist of three-dimensional brain MRI scans in NIfTI format. To harmonize structural characteristics across different data sources and accommodate model input requirements, all 3D volumes were converted into 2D axial slices. Given the brain's anatomical visibility and signal-to-noise ratio, slices from layers 80 to 120 were selected as the primary region of analysis, with systematic sampling every 5 layers to balance data representativeness and redundancy.

In line with the characteristics of the AD task, each slice was screened based on its corresponding tumor mask: slices with masks containing exclusively zero values were designated as “normal samples,” whereas slices containing non-zero mask pixels were treated as “anomalous samples” and retained along with their masks for use during the testing phase.

Data splitting adhered to strict sample independence. Normal slices were divided into training and testing sets at an 8:2 ratio, and the number of anomalous slices in the testing set was matched to the number of normal slices to construct a balanced test distribution. The final composition of each preprocessed dataset is as follows:

(1) \textbf{BraTS2018:} 1,082 normal slices for training, 274 regular slices for testing, 274 anomalous slices for testing with corresponding 274 segmentation masks.

(2) \textbf{MU-Glioma-Post:} 1,345 regular slices for training, 337 regular slices for testing, 337 anomalous slices for testing with corresponding 337 segmentation masks.

(3) \textbf{Pretreat-MetsToBrain-Masks:} 874 regular slices for training, 874 anomalous slices for testing with corresponding 874 segmentation masks.

\subsection{Comparison Experiments}

To comprehensively evaluate the effectiveness of the proposed model in AD and localization tasks, extensive comparative experiments were conducted on the BraTS2018 dataset against a variety of representative methods, including traditional reconstruction-based models such as STFPM \citep{wang2021student}, RD4AD \citep{deng2022anomaly}, AE-Flow \citep{zhao2023ae}, and DAE \citep{kascenas2023role}, as well as recently developed advanced approaches such as ReContrast \citep{guo2023recontrast}, GatingAno \citep{zhang2024anomaly}, skipTS \citep{liu2024anomaly}, ESCK-DRKD \citep{ge2025esc}, MMRAD \citep{li2025adapting}, and INP-Former \citep{luo2025exploring}.

\begin{table*}[htbp]
\centering
\caption{\normalfont\normalsize Comparison experiments results of different advanced models on the BraTS2018 dataset.}
\vspace{0.5em}
\label{table:competive_2018}
\resizebox{\textwidth}{!}{%
\begin{tabular}{@{}l l *{7}{c} c@{}}
\toprule
\textbf{Metrics} & \textbf{Variant} &
\circletextgray{F}\circletext{T1}\circletext{T2} &
\circletext{F}\circletextgray{T1}\circletext{T2} &
\circletext{F}\circletext{T1}\circletextgray{T2} &
\circletextgray{F}\circletextgray{T1}\circletext{T2} &
\circletextgray{F}\circletext{T1}\circletextgray{T2} &
\circletext{F}\circletextgray{T1}\circletextgray{T2} &
\circletextgray{F}\circletextgray{T1}\circletextgray{T2} &
\textbf{Avg} \\
\midrule

\multirow{11}{*}{\textbf{AUROC}}
  & \cellcolor{lightyellow} STFPM \citep{wang2021student}         & \cellcolor{lightyellow!25} 0.8208/0.9632 & \cellcolor{lightyellow!25} 0.6362/0.9294 & \cellcolor{lightyellow!25} 0.7571/0.9669 & \cellcolor{lightyellow!25} 0.7652/0.9479 & \cellcolor{lightyellow!25} 0.7194/0.9388 & \cellcolor{lightyellow!25} 0.6067/0.9320 & \cellcolor{lightyellow!25} 0.8393/0.9581 & \cellcolor{lightyellow!25} 0.7349/0.9480 \\
  & \cellcolor{lightyellow} RD4AD \citep{deng2022anomaly}        & \cellcolor{lightyellow!25} 0.9269/0.9783 & \cellcolor{lightyellow!25} 0.6833/0.9496 & \cellcolor{lightyellow!25} 0.8067/0.9728 & \cellcolor{lightyellow!25} 0.7282/0.9741 & \cellcolor{lightyellow!25} 0.8910/0.9771 & \cellcolor{lightyellow!25} 0.7764/0.9561 & \cellcolor{lightyellow!25} 0.8671/0.9742 & \cellcolor{lightyellow!25} 0.8114/0.9689 \\
  & \cellcolor{lightyellow} AE-Flow \citep{zhao2023ae}           & \cellcolor{lightyellow!25} 0.7359/0.8521 & \cellcolor{lightyellow!25} 0.6991/0.7516 & \cellcolor{lightyellow!25} 0.6636/0.8323 & \cellcolor{lightyellow!25} 0.7332/0.8846 & \cellcolor{lightyellow!25} 0.7609/0.8853 & \cellcolor{lightyellow!25} 0.7053/0.8969 & \cellcolor{lightyellow!25} 0.7637/0.8985 & \cellcolor{lightyellow!25} 0.7231/0.8573 \\
  & \cellcolor{lightyellow} DAE \citep{kascenas2023role}         & \cellcolor{lightyellow!25} 0.8499/0.9598 & \cellcolor{lightyellow!25} 0.6811/0.8726 & \cellcolor{lightyellow!25} 0.7953/0.9347 & \cellcolor{lightyellow!25} 0.8534/0.9690 & \cellcolor{lightyellow!25} 0.8404/0.9672 & \cellcolor{lightyellow!25} 0.7801/0.9293 & \cellcolor{lightyellow!25} 0.8604/0.9577 & \cellcolor{lightyellow!25} 0.8087/0.9415 \\
  & \cellcolor{lightyellow} ReContrast \citep{guo2023recontrast}& \cellcolor{lightyellow!25} 0.8924/0.9662 & \cellcolor{lightyellow!25} 0.7657/0.9524 & \cellcolor{lightyellow!25} 0.8317/0.9741 & \cellcolor{lightyellow!25} 0.9233/0.9830 & \cellcolor{lightyellow!25} 0.9188/0.9865 & \cellcolor{lightyellow!25} 0.8655/0.9792 & \cellcolor{lightyellow!25} 0.9293/0.9888 & \cellcolor{lightyellow!25} 0.8752/0.9757 \\
  & \cellcolor{lightyellow} GatingAno \citep{zhang2024anomaly}   & \cellcolor{lightyellow!25} 0.6851/0.8293 & \cellcolor{lightyellow!25} 0.6626/0.4730 & \cellcolor{lightyellow!25} 0.6516/0.9318 & \cellcolor{lightyellow!25} 0.6855/0.8621 & \cellcolor{lightyellow!25} 0.6694/0.8540 & \cellcolor{lightyellow!25} 0.6840/0.6640 & \cellcolor{lightyellow!25} 0.7012/0.8914 & \cellcolor{lightyellow!25} 0.6771/0.7865 \\
  & \cellcolor{lightyellow} skipTS \citep{liu2024anomaly}        & \cellcolor{lightyellow!25} 0.9213/0.9826 & \cellcolor{lightyellow!25} 0.7503/0.9599 & \cellcolor{lightyellow!25} 0.8194/0.9764 & \cellcolor{lightyellow!25} 0.7857/0.9644 & \cellcolor{lightyellow!25} 0.9173/0.9830 & \cellcolor{lightyellow!25} 0.7654/0.9571 & \cellcolor{lightyellow!25} 0.8926/0.9780 & \cellcolor{lightyellow!25} 0.8360/0.9716 \\
  & \cellcolor{lightyellow} ESCK-DRKD \citep{ge2025esc}          & \cellcolor{lightyellow!25} 0.7649/0.8695 & \cellcolor{lightyellow!25} 0.6838/0.9019 & \cellcolor{lightyellow!25} 0.6854/0.9093 & \cellcolor{lightyellow!25} 0.7871/0.8389 & \cellcolor{lightyellow!25} 0.7644/0.8439 & \cellcolor{lightyellow!25} 0.7457/0.8423 & \cellcolor{lightyellow!25} 0.7900/0.8889 & \cellcolor{lightyellow!25} 0.7459/0.8707 \\
  & \cellcolor{lightyellow} MMRAD \citep{li2025adapting}         & \cellcolor{lightyellow!25} 0.7306/0.8845 & \cellcolor{lightyellow!25} 0.7034/0.8121 & \cellcolor{lightyellow!25} 0.7104/0.8319 & \cellcolor{lightyellow!25} 0.7333/0.8939 & \cellcolor{lightyellow!25} 0.7232/0.8866 & \cellcolor{lightyellow!25} 0.7277/0.8649 & \cellcolor{lightyellow!25} 0.7726/0.9149 & \cellcolor{lightyellow!25} 0.7287/0.8698 \\
   & \cellcolor{lightyellow} INPFormer \citep{luo2025exploring}   
  & \cellcolor{lightyellow!25} 0.9383/0.9901 
 & \cellcolor{lightyellow!25} 0.9363/0.9877 
  & \cellcolor{lightyellow!25} 0.9060/0.9837 
  & \cellcolor{lightyellow!25} 0.9426/0.9890 
  & \cellcolor{lightyellow!25} 0.9452/0.9903 
  & \cellcolor{lightyellow!25} 0.9473/0.9903 
  & \cellcolor{lightyellow!25} 0.9475/0.9911 
  & \cellcolor{lightyellow!25} 0.9376/0.9889 \\

  & \cellcolor{lightyellow} Ours                                & \cellcolor{lightyellow!25} \textcolor{red}{0.9472}/\textcolor{red}{0.9921} & \cellcolor{lightyellow!25} \textcolor{red}{0.9458}/\textcolor{red}{0.9939} & \cellcolor{lightyellow!25} \textcolor{red}{0.9479}/\textcolor{red}{0.9921} & \cellcolor{lightyellow!25} \textcolor{red}{0.9462}/\textcolor{red}{0.9920} & \cellcolor{lightyellow!25} \textcolor{red}{0.9478}/\textcolor{red}{0.9921} & \cellcolor{lightyellow!25} \textcolor{red}{0.9481}/\textcolor{red}{0.9921} & \cellcolor{lightyellow!25} \textcolor{red}{0.9482}/\textcolor{red}{0.9921} & \cellcolor{lightyellow!25} \textcolor{red}{0.9473}/\textcolor{red}{0.9922} \\
\hdashline
\multirow{11}{*}{\textbf{AP}}
  & \cellcolor{lightred} STFPM \citep{wang2021student}           & \cellcolor{lightred!25} 0.7927/0.2642 & \cellcolor{lightred!25} 0.6054/0.1181 & \cellcolor{lightred!25} 0.7179/0.2928 & \cellcolor{lightred!25} 0.7197/0.1936 & \cellcolor{lightred!25} 0.6808/0.1453 & \cellcolor{lightred!25} 0.6194/0.1269 & \cellcolor{lightred!25} 0.7680/0.2525 & \cellcolor{lightred!25} 0.7006/0.1848 \\
  & \cellcolor{lightred} RD4AD \citep{deng2022anomaly}           & \cellcolor{lightred!25} 0.9274/0.4987 & \cellcolor{lightred!25} 0.6720/0.2283 & \cellcolor{lightred!25} 0.8141/0.4295 & \cellcolor{lightred!25} 0.7035/0.4372 & \cellcolor{lightred!25} 0.8773/0.4415 & \cellcolor{lightred!25} 0.7445/0.2766 & \cellcolor{lightred!25} 0.8493/0.4138 & \cellcolor{lightred!25} 0.7983/0.3894 \\
  & \cellcolor{lightred} AE-Flow \citep{zhao2023ae}              & \cellcolor{lightred!25} 0.7408/0.0518 & \cellcolor{lightred!25} 0.7261/0.0285 & \cellcolor{lightred!25} 0.6636/0.0523 & \cellcolor{lightred!25} 0.7031/0.0625 & \cellcolor{lightred!25} 0.7246/0.0608 & \cellcolor{lightred!25} 0.6796/0.0701 & \cellcolor{lightred!25} 0.7516/0.0950 & \cellcolor{lightred!25} 0.7128/0.0601 \\
  & \cellcolor{lightred} DAE \citep{kascenas2023role}            & \cellcolor{lightred!25} 0.8599/0.4158 & \cellcolor{lightred!25} 0.6823/0.1120 & \cellcolor{lightred!25} 0.7865/0.2456 & \cellcolor{lightred!25} 0.8661/0.3573 & \cellcolor{lightred!25} 0.8383/0.3246 & \cellcolor{lightred!25} 0.7693/0.2252 & \cellcolor{lightred!25} 0.8783/0.2714 & \cellcolor{lightred!25} 0.8115/0.2788 \\
  & \cellcolor{lightred} ReContrast \citep{guo2023recontrast}    & \cellcolor{lightred!25} 0.8945/0.4078 & \cellcolor{lightred!25} 0.7529/0.3109 & \cellcolor{lightred!25} 0.8116/0.4856 & \cellcolor{lightred!25} 0.8989/0.5905 & \cellcolor{lightred!25} 0.9118/0.6624 & \cellcolor{lightred!25} 0.8719/0.5715 & \cellcolor{lightred!25} 0.9323/0.7296 & \cellcolor{lightred!25} 0.8677/0.5369 \\
  & \cellcolor{lightred} GatingAno \citep{zhang2024anomaly}      & \cellcolor{lightred!25} 0.6895/0.0598 & \cellcolor{lightred!25} 0.7159/0.0442 & \cellcolor{lightred!25} 0.6634/0.1131 & \cellcolor{lightred!25} 0.7097/0.0777 & \cellcolor{lightred!25} 0.7093/0.0802 & \cellcolor{lightred!25} 0.6773/0.0375 & \cellcolor{lightred!25} 0.7316/0.0594 & \cellcolor{lightred!25} 0.6995/0.0674 \\
  & \cellcolor{lightred} skipTS \citep{liu2024anomaly}           & \cellcolor{lightred!25} 0.9025/0.4664 & \cellcolor{lightred!25} 0.7224/0.2746 & \cellcolor{lightred!25} 0.7826/0.3844 & \cellcolor{lightred!25} 0.7652/0.3256 & \cellcolor{lightred!25} 0.9081/0.4981 & \cellcolor{lightred!25} 0.7530/0.2880 & \cellcolor{lightred!25} 0.8749/0.4442 & \cellcolor{lightred!25} 0.8155/0.3830 \\
  & \cellcolor{lightred} ESCK-DRKD \citep{ge2025esc}             & \cellcolor{lightred!25} 0.7216/0.0538 & \cellcolor{lightred!25} 0.6930/0.0815 & \cellcolor{lightred!25} 0.6858/0.0903 & \cellcolor{lightred!25} 0.7630/0.0419 & \cellcolor{lightred!25} 0.7491/0.0425 & \cellcolor{lightred!25} 0.7205/0.0420 & \cellcolor{lightred!25} 0.7722/0.0635 & \cellcolor{lightred!25} 0.7293/0.0594 \\
  & \cellcolor{lightred} MMRAD \citep{li2025adapting}            & \cellcolor{lightred!25} 0.7506/0.0518 & \cellcolor{lightred!25} 0.7047/0.2139 & \cellcolor{lightred!25} 0.7218/0.0405 & \cellcolor{lightred!25} 0.7646/0.0618 & \cellcolor{lightred!25} 0.7470/0.0716 & \cellcolor{lightred!25} 0.7473/0.0747 & \cellcolor{lightred!25} 0.7967/0.0791 & \cellcolor{lightred!25} 0.7475/0.0848 \\
  & \cellcolor{lightred} INPFormer \citep{luo2025exploring}      & \cellcolor{lightred!25} 0.9369/0.7609 & \cellcolor{lightred!25} 0.9336/0.7323 & \cellcolor{lightred!25} 0.9092/0.6712 & \cellcolor{lightred!25} 0.9413/0.7641 & \cellcolor{lightred!25} 0.9424/0.7641 & \cellcolor{lightred!25} 0.9435/0.7660 & \cellcolor{lightred!25} 0.9436/0.7886 & \cellcolor{lightred!25} 0.9358/0.7510 \\
  & \cellcolor{lightred} Ours                                  
  & \cellcolor{lightred!25} \textcolor{red}{0.9438}/\textcolor{red}{0.7898} 
  & \cellcolor{lightred!25} \textcolor{red}{0.9429}/\textcolor{red}{0.7872} 
  & \cellcolor{lightred!25} \textcolor{red}{0.9468}/\textcolor{red}{0.7860} 
  & \cellcolor{lightred!25} \textcolor{red}{0.9433}/\textcolor{red}{0.7886} 
  & \cellcolor{lightred!25} \textcolor{red}{0.9469}/\textcolor{red}{0.7886} 
  & \cellcolor{lightred!25} \textcolor{red}{0.9470}/\textcolor{red}{0.7880} 
  & \cellcolor{lightred!25} \textcolor{red}{0.9472}/\textcolor{red}{0.7883} 
  & \cellcolor{lightred!25} \textcolor{red}{0.9454}/\textcolor{red}{0.7881} \\

\hdashline
\multirow{11}{*}{\textbf{F1}}
  & \cellcolor{lightpurple} STFPM \citep{wang2021student}         & \cellcolor{lightpurple!25} 0.7763/0.3530 & \cellcolor{lightpurple!25} 0.6835/0.1965 & \cellcolor{lightpurple!25} 0.7267/0.3737 & \cellcolor{lightpurple!25} 0.7474/0.2806 & \cellcolor{lightpurple!25} 0.7351/0.2285 & \cellcolor{lightpurple!25} 0.6708/0.1744 & \cellcolor{lightpurple!25} 0.8122/0.3383 & \cellcolor{lightpurple!25} 0.7360/0.2779 \\
  & \cellcolor{lightpurple} RD4AD \citep{deng2022anomaly}        & \cellcolor{lightpurple!25} 0.6667/0.1646 & \cellcolor{lightpurple!25} 0.6774/0.2652 & \cellcolor{lightpurple!25} 0.7161/0.4701 & \cellcolor{lightpurple!25} 0.6850/0.3021 & \cellcolor{lightpurple!25} 0.7690/0.4884 & \cellcolor{lightpurple!25} 0.7353/0.3419 & \cellcolor{lightpurple!25} 0.7253/0.4316 & \cellcolor{lightpurple!25} 0.7107/0.3520 \\
  & \cellcolor{lightpurple} AE-Flow \citep{zhao2023ae}           & \cellcolor{lightpurple!25} 0.7030/0.1025 & \cellcolor{lightpurple!25} 0.7152/0.0714 & \cellcolor{lightpurple!25} 0.7034/0.0974 & \cellcolor{lightpurple!25} 0.7166/0.1188 & \cellcolor{lightpurple!25} 0.7396/0.1180 & \cellcolor{lightpurple!25} 0.7163/0.1328 & \cellcolor{lightpurple!25} 0.7412/0.1345 & \cellcolor{lightpurple!25} 0.7193/0.1108 \\
  & \cellcolor{lightpurple} DAE \citep{kascenas2023role}         & \cellcolor{lightpurple!25} 0.7397/0.4560 & \cellcolor{lightpurple!25} 0.6944/0.1616 & \cellcolor{lightpurple!25} 0.7536/0.3225 & \cellcolor{lightpurple!25} 0.7734/0.4243 & \cellcolor{lightpurple!25} 0.7793/0.4079 & \cellcolor{lightpurple!25} 0.7296/0.3040 & \cellcolor{lightpurple!25} 0.7835/0.3590 & \cellcolor{lightpurple!25} 0.7505/0.3479 \\
  & \cellcolor{lightpurple} ReContrast \citep{guo2023recontrast}& \cellcolor{lightpurple!25} 0.8270/0.4457 & \cellcolor{lightpurple!25} 0.7221/0.3814 & \cellcolor{lightpurple!25} 0.7829/0.5104 & \cellcolor{lightpurple!25} 0.8702/0.6101 & \cellcolor{lightpurple!25} 0.8557/0.6432 & \cellcolor{lightpurple!25} 0.8015/0.5856 & \cellcolor{lightpurple!25} 0.8785/0.8572 & \cellcolor{lightpurple!25} 0.8197/0.5762 \\
  & \cellcolor{lightpurple} GatingAno \citep{zhang2024anomaly}   & \cellcolor{lightpurple!25} 0.7119/0.1145 & \cellcolor{lightpurple!25} 0.6959/0.0938 & \cellcolor{lightpurple!25} 0.7082/0.2028 & \cellcolor{lightpurple!25} 0.7062/0.1489 & \cellcolor{lightpurple!25} 0.7235/0.1474 & \cellcolor{lightpurple!25} 0.7314/0.0969 & \cellcolor{lightpurple!25} 0.7088/0.1101 & \cellcolor{lightpurple!25} 0.7123/0.1306 \\
  & \cellcolor{lightpurple} skipTS \citep{liu2024anomaly}        & \cellcolor{lightpurple!25} 0.9118/0.5126 & \cellcolor{lightpurple!25} 0.7361/0.3580 & \cellcolor{lightpurple!25} 0.8006/0.4614 & \cellcolor{lightpurple!25} 0.7753/0.3826 & \cellcolor{lightpurple!25} 0.9127/0.5295 & \cellcolor{lightpurple!25} 0.7591/0.3578 & \cellcolor{lightpurple!25} 0.8837/0.4876 & \cellcolor{lightpurple!25} 0.8256/0.4414 \\
  & \cellcolor{lightpurple} ESCK-DRKD \citep{ge2025esc}          & \cellcolor{lightpurple!25} 0.7572/0.1050 & \cellcolor{lightpurple!25} 0.6744/0.1672 & \cellcolor{lightpurple!25} 0.6831/0.1808 & \cellcolor{lightpurple!25} 0.7504/0.0938 & \cellcolor{lightpurple!25} 0.7405/0.0952 & \cellcolor{lightpurple!25} 0.7329/0.0958 & \cellcolor{lightpurple!25} 0.7542/0.1339 & \cellcolor{lightpurple!25} 0.7275/0.1245 \\
  & \cellcolor{lightpurple} MMRAD \citep{li2025adapting}         & \cellcolor{lightpurple!25} 0.7035/0.1028 & \cellcolor{lightpurple!25} 0.6851/0.2902 & \cellcolor{lightpurple!25} 0.6826/0.0952 & \cellcolor{lightpurple!25} 0.7077/0.1380 & \cellcolor{lightpurple!25} 0.6890/0.1519 & \cellcolor{lightpurple!25} 0.6817/0.1525 & \cellcolor{lightpurple!25} 0.7156/0.1548 & \cellcolor{lightpurple!25} 0.6950/0.1551 \\
  & \cellcolor{lightpurple} INPFormer \citep{luo2025exploring}   & \cellcolor{lightpurple!25} 0.8736/0.7151 & \cellcolor{lightpurple!25} 0.8644/0.6877 & \cellcolor{lightpurple!25} 0.8298/0.6391 & \cellcolor{lightpurple!25} 0.8730/0.7157 & \cellcolor{lightpurple!25} 0.8769/0.7136 & \cellcolor{lightpurple!25} 0.8851/0.7230 & \cellcolor{lightpurple!25} 0.8687/0.6950 & \cellcolor{lightpurple!25} 0.8674/0.6985 \\
  & \cellcolor{lightpurple} Ours                                
  & \cellcolor{lightpurple!25} \textcolor{red}{0.8852}/\textcolor{red}{0.7359}
  & \cellcolor{lightpurple!25} \textcolor{red}{0.8849}/\textcolor{red}{0.7330}
  & \cellcolor{lightpurple!25} \textcolor{red}{0.8841}/\textcolor{red}{0.7344}
  & \cellcolor{lightpurple!25} \textcolor{red}{0.8856}/\textcolor{red}{0.7352}
  & \cellcolor{lightpurple!25} \textcolor{red}{0.8877}/\textcolor{red}{0.7353}
  & \cellcolor{lightpurple!25} \textcolor{red}{0.8881}/\textcolor{red}{0.7349}
  & \cellcolor{lightpurple!25} \textcolor{red}{0.8881}/\textcolor{red}{0.7369}
  & \cellcolor{lightpurple!25} \textcolor{red}{0.8862}/\textcolor{red}{0.7351} \\

\hdashline

\multirow{11}{*}{\textbf{AUPRO}}
  & \cellcolor{darkblue} STFPM \citep{wang2021student}           & \cellcolor{darkblue!25} 0.8293 & \cellcolor{darkblue!25} 0.7051 & \cellcolor{darkblue!25} 0.7815 & \cellcolor{darkblue!25} 0.7622 & \cellcolor{darkblue!25} 0.7178 & \cellcolor{darkblue!25} 0.7223 & \cellcolor{darkblue!25} 0.7929 & \cellcolor{darkblue!25} 0.7587 \\
  & \cellcolor{darkblue} RD4AD \citep{deng2022anomaly}          & \cellcolor{darkblue!25} 0.7905 & \cellcolor{darkblue!25} 0.6948 & \cellcolor{darkblue!25} 0.7620 & \cellcolor{darkblue!25} 0.7210 & \cellcolor{darkblue!25} 0.7685 & \cellcolor{darkblue!25} 0.7212 & \cellcolor{darkblue!25} 0.7800 & \cellcolor{darkblue!25} 0.7483 \\
  & \cellcolor{darkblue} AE-Flow \citep{zhao2023ae}             & \cellcolor{darkblue!25} 0.5496 & \cellcolor{darkblue!25} 0.3030 & \cellcolor{darkblue!25} 0.4615 & \cellcolor{darkblue!25} 0.5927 & \cellcolor{darkblue!25} 0.5585 & \cellcolor{darkblue!25} 0.5923 & \cellcolor{darkblue!25} 0.5964 & \cellcolor{darkblue!25} 0.5220 \\
  & \cellcolor{darkblue} DAE \citep{kascenas2023role}           & \cellcolor{darkblue!25} 0.5756 & \cellcolor{darkblue!25} 0.5206 & \cellcolor{darkblue!25} 0.5592 & \cellcolor{darkblue!25} 0.5812 & \cellcolor{darkblue!25} 0.5801 & \cellcolor{darkblue!25} 0.5563 & \cellcolor{darkblue!25} 0.5744 & \cellcolor{darkblue!25} 0.5639 \\
  & \cellcolor{darkblue} ReContrast \citep{guo2023recontrast}   & \cellcolor{darkblue!25} 0.8456 & \cellcolor{darkblue!25} 0.7232 & \cellcolor{darkblue!25} 0.7715 & \cellcolor{darkblue!25} 0.8426 & \cellcolor{darkblue!25} 0.8628 & \cellcolor{darkblue!25} 0.7832 & \cellcolor{darkblue!25} 0.8572 & \cellcolor{darkblue!25} 0.8123 \\
  & \cellcolor{darkblue} GatingAno \citep{zhang2024anomaly}     & \cellcolor{darkblue!25} 0.4510 & \cellcolor{darkblue!25} 0.4810 & \cellcolor{darkblue!25} 0.7178 & \cellcolor{darkblue!25} 0.6493 & \cellcolor{darkblue!25} 0.6426 & \cellcolor{darkblue!25} 0.3444 & \cellcolor{darkblue!25} 0.5329 & \cellcolor{darkblue!25} 0.5456 \\
  & \cellcolor{darkblue} skipTS \citep{liu2024anomaly}          & \cellcolor{darkblue!25} 0.7921 & \cellcolor{darkblue!25} 0.6955 & \cellcolor{darkblue!25} 0.7397 & \cellcolor{darkblue!25} 0.7014 & \cellcolor{darkblue!25} 0.7834 & \cellcolor{darkblue!25} 0.6737 & \cellcolor{darkblue!25} 0.7565 & \cellcolor{darkblue!25} 0.7346 \\
  & \cellcolor{darkblue} ESCK-DRKD \citep{ge2025esc}            & \cellcolor{darkblue!25} 0.5648 & \cellcolor{darkblue!25} 0.5858 & \cellcolor{darkblue!25} 0.5907 & \cellcolor{darkblue!25} 0.5446 & \cellcolor{darkblue!25} 0.5481 & \cellcolor{darkblue!25} 0.5471 & \cellcolor{darkblue!25} 0.5772 & \cellcolor{darkblue!25} 0.5655 \\
  & \cellcolor{darkblue} MMRAD \citep{li2025adapting}           & \cellcolor{darkblue!25} 0.5386 & \cellcolor{darkblue!25} 0.5016 & \cellcolor{darkblue!25} 0.5026 & \cellcolor{darkblue!25} 0.5532 & \cellcolor{darkblue!25} 0.5374 & \cellcolor{darkblue!25} 0.5463 & \cellcolor{darkblue!25} 0.5750 & \cellcolor{darkblue!25} 0.5364 \\
  & \cellcolor{darkblue} INPFormer \citep{luo2025exploring}     & \cellcolor{darkblue!25} 0.8555 & \cellcolor{darkblue!25} 0.8521 & \cellcolor{darkblue!25} 0.8106 & \cellcolor{darkblue!25} 0.8531 & \cellcolor{darkblue!25} 0.8596 & \cellcolor{darkblue!25} 0.8622 & \cellcolor{darkblue!25} 0.8576 & \cellcolor{darkblue!25} 0.8501 \\
  & \cellcolor{darkblue} Ours                                  
  & \cellcolor{darkblue!25} \textcolor{red}{0.8749}
  & \cellcolor{darkblue!25} \textcolor{red}{0.8768}
  & \cellcolor{darkblue!25} \textcolor{red}{0.8748}
  & \cellcolor{darkblue!25} \textcolor{red}{0.8757}
  & \cellcolor{darkblue!25} \textcolor{red}{0.8746}
  & \cellcolor{darkblue!25} \textcolor{red}{0.8752}
  & \cellcolor{darkblue!25} \textcolor{red}{0.8796}
  & \cellcolor{darkblue!25} \textcolor{red}{0.8760} \\

\bottomrule
\end{tabular}
}
\end{table*}

\begin{table*}[htbp]
\centering
\caption{\normalfont\normalsize Comparison results experiments of different advanced models on the MU-Glioma-Post dataset.}
\vspace{0.5em}
\label{table:competive_mu}
\resizebox{\textwidth}{!}{%
\begin{tabular}{@{}l l *{7}{c} c@{}}
\toprule
\textbf{Metrics} & \textbf{Variant} &
\circletextgray{F}\circletext{T1}\circletext{T2} &
\circletext{F}\circletextgray{T1}\circletext{T2} &
\circletext{F}\circletext{T1}\circletextgray{T2} &
\circletextgray{F}\circletextgray{T1}\circletext{T2} &
\circletextgray{F}\circletext{T1}\circletextgray{T2} &
\circletext{F}\circletextgray{T1}\circletextgray{T2} &
\circletextgray{F}\circletextgray{T1}\circletextgray{T2} &
\textbf{Avg} \\
\midrule

\multirow{11}{*}{\textbf{AUROC}}
  & \cellcolor{lightyellow} STFPM \citep{wang2021student}        & \cellcolor{lightyellow!25} 0.7249/0.9626 & \cellcolor{lightyellow!25} 0.6351/0.9562 & \cellcolor{lightyellow!25} 0.7035/0.9690 & \cellcolor{lightyellow!25} 0.6827/0.9459 & \cellcolor{lightyellow!25} 0.7989/0.9677 & \cellcolor{lightyellow!25} 0.6067/0.9320 & \cellcolor{lightyellow!25} 0.7165/0.9499 & \cellcolor{lightyellow!25} 0.6955/0.9548 \\
  & \cellcolor{lightyellow} RD4AD \citep{deng2022anomaly}        & \cellcolor{lightyellow!25} 0.7767/0.9659 & \cellcolor{lightyellow!25} 0.7052/0.9342 & \cellcolor{lightyellow!25} 0.7965/0.9760 & \cellcolor{lightyellow!25} 0.6961/0.9436 & \cellcolor{lightyellow!25} 0.8210/0.9793 & \cellcolor{lightyellow!25} 0.6271/0.9393 & \cellcolor{lightyellow!25} 0.7416/0.9695 & \cellcolor{lightyellow!25} 0.7377/0.9583 \\
  & \cellcolor{lightyellow} AE-Flow \citep{zhao2023ae}          & \cellcolor{lightyellow!25} 0.7439/0.8352 & \cellcolor{lightyellow!25} 0.7001/0.7985 & \cellcolor{lightyellow!25} 0.7283/0.8580 & \cellcolor{lightyellow!25} 0.6638/0.8071 & \cellcolor{lightyellow!25} 0.7254/0.8601 & \cellcolor{lightyellow!25} 0.6811/0.8131 & \cellcolor{lightyellow!25} 0.7690/0.8961 & \cellcolor{lightyellow!25} 0.7159/0.8383 \\
  & \cellcolor{lightyellow} DAE \citep{kascenas2023role}        & \cellcolor{lightyellow!25} 0.8212/0.8591 & \cellcolor{lightyellow!25} 0.7476/0.8986 & \cellcolor{lightyellow!25} 0.7341/0.9167 & \cellcolor{lightyellow!25} 0.7649/0.8921 & \cellcolor{lightyellow!25} 0.8058/0.9299 & \cellcolor{lightyellow!25} 0.7807/0.9401 & \cellcolor{lightyellow!25} 0.8535/0.9487 & \cellcolor{lightyellow!25} 0.7868/0.9122 \\
  & \cellcolor{lightyellow} ReContrast \citep{guo2023recontrast}& \cellcolor{lightyellow!25} 0.8537/0.9751 & \cellcolor{lightyellow!25} 0.7671/0.9707 & \cellcolor{lightyellow!25} 0.8052/0.9776 & \cellcolor{lightyellow!25} 0.8564/0.9854 & \cellcolor{lightyellow!25} 0.8693/0.9888 & \cellcolor{lightyellow!25} 0.8147/0.9833 & \cellcolor{lightyellow!25} 0.8720/0.9892 & \cellcolor{lightyellow!25} 0.8341/0.9814 \\
  & \cellcolor{lightyellow} GatingAno \citep{zhang2024anomaly}  & \cellcolor{lightyellow!25} 0.7703/0.8956 & \cellcolor{lightyellow!25} 0.7224/0.8277 & \cellcolor{lightyellow!25} 0.7833/0.9011 & \cellcolor{lightyellow!25} 0.7510/0.8607 & \cellcolor{lightyellow!25} 0.7512/0.8692 & \cellcolor{lightyellow!25} 0.7784/0.8935 & \cellcolor{lightyellow!25} 0.7849/0.9179 & \cellcolor{lightyellow!25} 0.7631/0.8808 \\
  & \cellcolor{lightyellow} skipTS \citep{liu2024anomaly}       & \cellcolor{lightyellow!25} 0.7816/0.9749 & \cellcolor{lightyellow!25} 0.6165/0.9547 & \cellcolor{lightyellow!25} 0.8171/0.9768 & \cellcolor{lightyellow!25} 0.6487/0.9554 & \cellcolor{lightyellow!25} 0.7465/0.9770 & \cellcolor{lightyellow!25} 0.6676/0.9450 & \cellcolor{lightyellow!25} 0.6707/0.9610 & \cellcolor{lightyellow!25} 0.7069/0.9635 \\
  & \cellcolor{lightyellow} ESCK-DRKD \citep{ge2025esc}         & \cellcolor{lightyellow!25} 0.7718/0.8734 & \cellcolor{lightyellow!25} 0.6738/0.7998 & \cellcolor{lightyellow!25} 0.6423/0.7630 & \cellcolor{lightyellow!25} 0.7239/0.8232 & \cellcolor{lightyellow!25} 0.7044/0.8044 & \cellcolor{lightyellow!25} 0.6738/0.7998 & \cellcolor{lightyellow!25} 0.8026/0.8866 & \cellcolor{lightyellow!25} 0.7132/0.8215 \\
  & \cellcolor{lightyellow} MMRAD \citep{li2025adapting}        & \cellcolor{lightyellow!25} 0.7289/0.8992 & \cellcolor{lightyellow!25} 0.7186/0.8876 & \cellcolor{lightyellow!25} 0.6918/0.8761 & \cellcolor{lightyellow!25} 0.7096/0.8811 & \cellcolor{lightyellow!25} 0.7010/0.8891 & \cellcolor{lightyellow!25} 0.7087/0.8832 & \cellcolor{lightyellow!25} 0.7297/0.9003 & \cellcolor{lightyellow!25} 0.7126/0.8881 \\
  & \cellcolor{lightyellow} INPFormer \citep{luo2025exploring}  & \cellcolor{lightyellow!25} 0.8189/0.9787 & \cellcolor{lightyellow!25} 0.7703/0.9758 & \cellcolor{lightyellow!25} 0.7956/0.9782 & \cellcolor{lightyellow!25} 0.8248/0.9797 & \cellcolor{lightyellow!25} 0.8361/0.9804 & \cellcolor{lightyellow!25} 0.8413/0.9809 & \cellcolor{lightyellow!25} 0.8734/0.9845 & \cellcolor{lightyellow!25} 0.8229/0.9797 \\
  & \cellcolor{lightyellow} Ours
  & \cellcolor{lightyellow!25} \textcolor{red}{0.8901/0.9889}
  & \cellcolor{lightyellow!25} \textcolor{red}{0.8908/0.9891}
  & \cellcolor{lightyellow!25} \textcolor{red}{0.8902/0.9889}
  & \cellcolor{lightyellow!25} \textcolor{red}{0.8902/0.9889}
  & \cellcolor{lightyellow!25} \textcolor{red}{0.8907/0.9890}
  & \cellcolor{lightyellow!25} \textcolor{red}{0.8902/0.9889}
  & \cellcolor{lightyellow!25} \textcolor{red}{0.8908/0.9899}
  & \cellcolor{lightyellow!25} \textcolor{red}{0.8905/0.9891} \\

\hdashline

\multirow{11}{*}{\textbf{AP}}
  & \cellcolor{lightred} STFPM \citep{wang2021student}        & \cellcolor{lightred!25} 0.6995/0.2217 & \cellcolor{lightred!25} 0.5828/0.2099 & \cellcolor{lightred!25} 0.6737/0.2538 & \cellcolor{lightred!25} 0.6950/0.2002 & \cellcolor{lightred!25} 0.7989/0.2824 & \cellcolor{lightred!25} 0.6194/0.1269 & \cellcolor{lightred!25} 0.7181/0.1533 & \cellcolor{lightred!25} 0.6839/0.2069 \\
  & \cellcolor{lightred} RD4AD \citep{deng2022anomaly}        & \cellcolor{lightred!25} 0.7951/0.2918 & \cellcolor{lightred!25} 0.7480/0.1840 & \cellcolor{lightred!25} 0.7875/0.4086 & \cellcolor{lightred!25} 0.6358/0.2486 & \cellcolor{lightred!25} 0.8115/0.5069 & \cellcolor{lightred!25} 0.6363/0.1988 & \cellcolor{lightred!25} 0.7281/0.3586 & \cellcolor{lightred!25} 0.7346/0.3139 \\
  & \cellcolor{lightred} AE-Flow \citep{zhao2023ae}          & \cellcolor{lightred!25} 0.7499/0.0392 & \cellcolor{lightred!25} 0.7127/0.0294 & \cellcolor{lightred!25} 0.6333/0.0501 & \cellcolor{lightred!25} 0.7032/0.0645 & \cellcolor{lightred!25} 0.6338/0.0480 & \cellcolor{lightred!25} 0.6879/0.0761 & \cellcolor{lightred!25} 0.7600/0.0608 & \cellcolor{lightred!25} 0.6973/0.0526 \\
  & \cellcolor{lightred} DAE \citep{kascenas2023role}        & \cellcolor{lightred!25} 0.7776/0.0468 & \cellcolor{lightred!25} 0.7290/0.1044 & \cellcolor{lightred!25} 0.7524/0.1553 & \cellcolor{lightred!25} 0.7179/0.0693 & \cellcolor{lightred!25} 0.7772/0.1328 & \cellcolor{lightred!25} 0.7770/0.1803 & \cellcolor{lightred!25} 0.8266/0.2039 & \cellcolor{lightred!25} 0.7654/0.1275 \\
  & \cellcolor{lightred} ReContrast \citep{guo2023recontrast}& \cellcolor{lightred!25} 0.8576/0.4432 & \cellcolor{lightred!25} 0.7677/0.4415 & \cellcolor{lightred!25} 0.8205/0.5129 & \cellcolor{lightred!25} 0.8708/0.5874 & \cellcolor{lightred!25} 0.8813/0.6787 & \cellcolor{lightred!25} 0.8049/0.5480 & \cellcolor{lightred!25} 0.8880/0.7038 & \cellcolor{lightred!25} 0.8415/0.5594 \\
  & \cellcolor{lightred} GatingAno \citep{zhang2024anomaly}  & \cellcolor{lightred!25} 0.7674/0.1397 & \cellcolor{lightred!25} 0.7174/0.0613 & \cellcolor{lightred!25} 0.8022/0.1112 & \cellcolor{lightred!25} 0.7698/0.0779 & \cellcolor{lightred!25} 0.7639/0.0921 & \cellcolor{lightred!25} 0.7765/0.1595 & \cellcolor{lightred!25} 0.7872/0.0707 & \cellcolor{lightred!25} 0.7692/0.1018 \\
  & \cellcolor{lightred} skipTS \citep{liu2024anomaly}       & \cellcolor{lightred!25} 0.7162/0.2918 & \cellcolor{lightred!25} 0.6108/0.1840 & \cellcolor{lightred!25} 0.8192/0.4086 & \cellcolor{lightred!25} 0.6118/0.2486 & \cellcolor{lightred!25} 0.6710/0.5069 & \cellcolor{lightred!25} 0.6333/0.1988 & \cellcolor{lightred!25} 0.6410/0.3586 & \cellcolor{lightred!25} 0.6719/0.3139 \\
  & \cellcolor{lightred} ESCK-DRKD \citep{ge2025esc}         & \cellcolor{lightred!25} 0.7852/0.0500 & \cellcolor{lightred!25} 0.1692/0.0888 & \cellcolor{lightred!25} 0.6710/0.0776 & \cellcolor{lightred!25} 0.7042/0.0997 & \cellcolor{lightred!25} 0.6806/0.0904 & \cellcolor{lightred!25} 0.6921/0.0888 & \cellcolor{lightred!25} 0.7506/0.1193 & \cellcolor{lightred!25} 0.6361/0.0878 \\
  & \cellcolor{lightred} MMRAD \citep{li2025adapting}        & \cellcolor{lightred!25} 0.7469/0.0859 & \cellcolor{lightred!25} 0.7432/0.0742 & \cellcolor{lightred!25} 0.7009/0.0683 & \cellcolor{lightred!25} 0.7254/0.0699 & \cellcolor{lightred!25} 0.7234/0.0789 & \cellcolor{lightred!25} 0.7334/0.0772 & \cellcolor{lightred!25} 0.7527/0.0862 & \cellcolor{lightred!25} 0.7323/0.0772 \\
  & \cellcolor{lightred} INPFormer \citep{luo2025exploring}  & \cellcolor{lightred!25} 0.8653/0.5788 & \cellcolor{lightred!25} 0.8333/0.5055 & \cellcolor{lightred!25} 0.8479/0.5662 & \cellcolor{lightred!25} 0.8689/0.6115 & \cellcolor{lightred!25} 0.8770/0.6235 & \cellcolor{lightred!25} 0.8789/0.6378 & \cellcolor{lightred!25} 0.8915/0.6710 & \cellcolor{lightred!25} 0.8661/0.6006 \\
  & \cellcolor{lightred} Ours
  & \cellcolor{lightred!25} \textcolor{red}{0.9071/0.7188}
  & \cellcolor{lightred!25} \textcolor{red}{0.9071/0.7186}
  & \cellcolor{lightred!25} \textcolor{red}{0.9075/0.7195}
  & \cellcolor{lightred!25} \textcolor{red}{0.9069/0.7180}
  & \cellcolor{lightred!25} \textcolor{red}{0.9075/0.7199}
  & \cellcolor{lightred!25} \textcolor{red}{0.9067/0.7180}
  & \cellcolor{lightred!25} \textcolor{red}{0.9075/0.7201}
  & \cellcolor{lightred!25} \textcolor{red}{0.9072/0.7189} \\

\hdashline

\multirow{11}{*}{\textbf{F1}}
  & \cellcolor{lightpurple} STFPM \citep{wang2021student}        & \cellcolor{lightpurple!25} 0.7133/0.3155 & \cellcolor{lightpurple!25} 0.6834/0.2896 & \cellcolor{lightpurple!25} 0.6867/0.3443 & \cellcolor{lightpurple!25} 0.6860/0.2492 & \cellcolor{lightpurple!25} 0.7490/0.3599 & \cellcolor{lightpurple!25} 0.6708/0.1744 & \cellcolor{lightpurple!25} 0.6952/0.2527 & \cellcolor{lightpurple!25} 0.6978/0.2837 \\
  & \cellcolor{lightpurple} RD4AD \citep{deng2022anomaly}        & \cellcolor{lightpurple!25} 0.7107/0.3541 & \cellcolor{lightpurple!25} 0.6687/0.2290 & \cellcolor{lightpurple!25} 0.7494/0.4439 & \cellcolor{lightpurple!25} 0.6849/0.3005 & \cellcolor{lightpurple!25} 0.7485/0.5334 & \cellcolor{lightpurple!25} 0.6680/0.1988 & \cellcolor{lightpurple!25} 0.7082/0.4055 & \cellcolor{lightpurple!25} 0.7053/0.3522 \\
  & \cellcolor{lightpurple} AE-Flow \citep{zhao2023ae}          & \cellcolor{lightpurple!25} 0.6955/0.0857 & \cellcolor{lightpurple!25} 0.6835/0.0697 & \cellcolor{lightpurple!25} 0.6948/0.1072 & \cellcolor{lightpurple!25} 0.6902/0.1332 & \cellcolor{lightpurple!25} 0.7382/0.0942 & \cellcolor{lightpurple!25} 0.7024/0.1224 & \cellcolor{lightpurple!25} 0.7583/0.1454 & \cellcolor{lightpurple!25} 0.7090/0.1083 \\
  & \cellcolor{lightpurple} DAE \citep{kascenas2023role}        & \cellcolor{lightpurple!25} 0.7947/0.0986 & \cellcolor{lightpurple!25} 0.7320/0.1807 & \cellcolor{lightpurple!25} 0.6926/0.2430 & \cellcolor{lightpurple!25} 0.7512/0.1242 & \cellcolor{lightpurple!25} 0.7642/0.2056 & \cellcolor{lightpurple!25} 0.7280/0.2605 & \cellcolor{lightpurple!25} 0.7984/0.3000 & \cellcolor{lightpurple!25} 0.7516/0.2018 \\
  & \cellcolor{lightpurple} ReContrast \citep{guo2023recontrast}& \cellcolor{lightpurple!25} 0.8576/0.4432 & \cellcolor{lightpurple!25} 0.7677/0.4415 & \cellcolor{lightpurple!25} 0.8205/0.5129 & \cellcolor{lightpurple!25} 0.8708/0.5874 & \cellcolor{lightpurple!25} 0.8813/0.6787 & \cellcolor{lightpurple!25} 0.8049/0.5480 & \cellcolor{lightpurple!25} 0.8880/0.7038 & \cellcolor{lightpurple!25} 0.8415/0.5594 \\
  & \cellcolor{lightpurple} GatingAno \citep{zhang2024anomaly}  & \cellcolor{lightpurple!25} 0.7157/0.1503 & \cellcolor{lightpurple!25} 0.6988/0.1217 & \cellcolor{lightpurple!25} 0.7195/0.2018 & \cellcolor{lightpurple!25} 0.7023/0.1591 & \cellcolor{lightpurple!25} 0.6980/0.1834 & \cellcolor{lightpurple!25} 0.7370/0.2687 & \cellcolor{lightpurple!25} 0.7644/0.2545 & \cellcolor{lightpurple!25} 0.7194/0.1914 \\
  & \cellcolor{lightpurple} skipTS \citep{liu2024anomaly}       & \cellcolor{lightpurple!25} 0.7475/0.3541 & \cellcolor{lightpurple!25} 0.6136/0.2290 & \cellcolor{lightpurple!25} 0.8181/0.4439 & \cellcolor{lightpurple!25} 0.6297/0.3005 & \cellcolor{lightpurple!25} 0.7067/0.5334 & \cellcolor{lightpurple!25} 0.6500/0.1988 & \cellcolor{lightpurple!25} 0.6556/0.4055 & \cellcolor{lightpurple!25} 0.6887/0.3522 \\
  & \cellcolor{lightpurple} ESCK-DRKD \citep{ge2025esc}         & \cellcolor{lightpurple!25} 0.7560/0.1138 & \cellcolor{lightpurple!25} 0.6835/0.1627 & \cellcolor{lightpurple!25} 0.6667/0.1454 & \cellcolor{lightpurple!25} 0.7251/0.1916 & \cellcolor{lightpurple!25} 0.7062/0.1848 & \cellcolor{lightpurple!25} 0.6835/0.1627 & \cellcolor{lightpurple!25} 0.7596/0.4493 & \cellcolor{lightpurple!25} 0.7115/0.2015 \\
  & \cellcolor{lightpurple} MMRAD \citep{li2025adapting}        & \cellcolor{lightpurple!25} 0.7091/0.1583 & \cellcolor{lightpurple!25} 0.6985/0.1419 & \cellcolor{lightpurple!25} 0.6962/0.1313 & \cellcolor{lightpurple!25} 0.7000/0.1341 & \cellcolor{lightpurple!25} 0.6867/0.1548 & \cellcolor{lightpurple!25} 0.6953/0.1531 & \cellcolor{lightpurple!25} 0.7051/0.1656 & \cellcolor{lightpurple!25} 0.6987/0.1484 \\
  & \cellcolor{lightpurple} INPFormer \citep{luo2025exploring}  & \cellcolor{lightpurple!25} 0.7560/0.5429 & \cellcolor{lightpurple!25} 0.7255/0.4854 & \cellcolor{lightpurple!25} 0.7321/0.5349 & \cellcolor{lightpurple!25} 0.7687/0.5708 & \cellcolor{lightpurple!25} 0.7729/0.5815 & \cellcolor{lightpurple!25} 0.7742/0.5975 & \cellcolor{lightpurple!25} 0.7894/0.6334 & \cellcolor{lightpurple!25} 0.7598/0.5638 \\
  & \cellcolor{lightpurple} Ours
  & \cellcolor{lightpurple!25} \textcolor{red}{0.8017/0.6707}
  & \cellcolor{lightpurple!25} \textcolor{red}{0.8023/0.6701}
  & \cellcolor{lightpurple!25} \textcolor{red}{0.8021/0.6715}
  & \cellcolor{lightpurple!25} \textcolor{red}{0.8017/0.6698}
  & \cellcolor{lightpurple!25} \textcolor{red}{0.8012/0.6713}
  & \cellcolor{lightpurple!25} \textcolor{red}{0.8017/0.6708}
  & \cellcolor{lightpurple!25} \textcolor{red}{0.8035/0.6726}
  & \cellcolor{lightpurple!25} \textcolor{red}{0.8020/0.6710} \\

\hdashline

\multirow{11}{*}{\textbf{AUPRO}}
  & \cellcolor{darkblue} RD4AD \citep{deng2022anomaly}        & \cellcolor{darkblue!25} 0.8217 & \cellcolor{darkblue!25} 0.7249 & \cellcolor{darkblue!25} 0.8440 & \cellcolor{darkblue!25} 0.7737 & \cellcolor{darkblue!25} 0.8791 & \cellcolor{darkblue!25} 0.7690 & \cellcolor{darkblue!25} 0.8575 & \cellcolor{darkblue!25} 0.8100 \\
  & \cellcolor{darkblue} AE-Flow \citep{zhao2023ae}          & \cellcolor{darkblue!25} 0.4985 & \cellcolor{darkblue!25} 0.3389 & \cellcolor{darkblue!25} 0.5590 & \cellcolor{darkblue!25} 0.6463 & \cellcolor{darkblue!25} 0.4866 & \cellcolor{darkblue!25} 0.5980 & \cellcolor{darkblue!25} 0.6711 & \cellcolor{darkblue!25} 0.5426 \\
  & \cellcolor{darkblue} DAE \citep{kascenas2023role}        & \cellcolor{darkblue!25} 0.5118 & \cellcolor{darkblue!25} 0.5375 & \cellcolor{darkblue!25} 0.5497 & \cellcolor{darkblue!25} 0.5344 & \cellcolor{darkblue!25} 0.5574 & \cellcolor{darkblue!25} 0.5637 & \cellcolor{darkblue!25} 0.5689 & \cellcolor{darkblue!25} 0.5462 \\
  & \cellcolor{darkblue} ReContrast \citep{guo2023recontrast}& \cellcolor{darkblue!25} 0.8604 & \cellcolor{darkblue!25} 0.8005 & \cellcolor{darkblue!25} 0.8119 & \cellcolor{darkblue!25} 0.8707 & \cellcolor{darkblue!25} 0.8965 & \cellcolor{darkblue!25} 0.8515 & \cellcolor{darkblue!25} 0.8912 & \cellcolor{darkblue!25} 0.8547 \\
  & \cellcolor{darkblue} GatingAno \citep{zhang2024anomaly}  & \cellcolor{darkblue!25} 0.5316 & \cellcolor{darkblue!25} 0.5306 & \cellcolor{darkblue!25} 0.5728 & \cellcolor{darkblue!25} 0.6362 & \cellcolor{darkblue!25} 0.5899 & \cellcolor{darkblue!25} 0.6150 & \cellcolor{darkblue!25} 0.6733 & \cellcolor{darkblue!25} 0.5928 \\
  & \cellcolor{darkblue} skipTS \citep{liu2024anomaly}       & \cellcolor{darkblue!25} 0.8534 & \cellcolor{darkblue!25} 0.7581 & \cellcolor{darkblue!25} 0.8541 & \cellcolor{darkblue!25} 0.7811 & \cellcolor{darkblue!25} 0.8597 & \cellcolor{darkblue!25} 0.7608 & \cellcolor{darkblue!25} 0.8039 & \cellcolor{darkblue!25} 0.8102 \\
  & \cellcolor{darkblue} ESCK-DRKD \citep{ge2025esc}         & \cellcolor{darkblue!25} 0.5313 & \cellcolor{darkblue!25} 0.5398 & \cellcolor{darkblue!25} 0.5357 & \cellcolor{darkblue!25} 0.5536 & \cellcolor{darkblue!25} 0.5398 & \cellcolor{darkblue!25} 0.5483 & \cellcolor{darkblue!25} 0.5838 & \cellcolor{darkblue!25} 0.5475 \\
  & \cellcolor{darkblue} MMRAD \citep{li2025adapting}        & \cellcolor{darkblue!25} 0.6228 & \cellcolor{darkblue!25} 0.5815 & \cellcolor{darkblue!25} 0.5440 & \cellcolor{darkblue!25} 0.5602 & \cellcolor{darkblue!25} 0.5909 & \cellcolor{darkblue!25} 0.5749 & \cellcolor{darkblue!25} 0.6217 & \cellcolor{darkblue!25} 0.5851 \\
  & \cellcolor{darkblue} INPFormer \citep{luo2025exploring}  & \cellcolor{darkblue!25} 0.8397 & \cellcolor{darkblue!25} 0.8291 & \cellcolor{darkblue!25} 0.8370 & \cellcolor{darkblue!25} 0.8439 & \cellcolor{darkblue!25} 0.8469 & \cellcolor{darkblue!25} 0.8487 & \cellcolor{darkblue!25} 0.8729 & \cellcolor{darkblue!25} 0.8455 \\
   & \cellcolor{darkblue} Ours
  & \cellcolor{darkblue!25} \textcolor{red}{0.8977}
  & \cellcolor{darkblue!25} \textcolor{red}{0.9007}
  & \cellcolor{darkblue!25} \textcolor{red}{0.8973}
  & \cellcolor{darkblue!25} \textcolor{red}{0.8987}
  & \cellcolor{darkblue!25} \textcolor{red}{0.8978}
  & \cellcolor{darkblue!25} \textcolor{red}{0.8967}
  & \cellcolor{darkblue!25} \textcolor{red}{0.9012}
  & \cellcolor{darkblue!25} \textcolor{red}{0.8986} \\

\bottomrule
\end{tabular}%
}
\end{table*}

As shown in Table \ref{table:competive_2018}, Table \ref{table:competive_mu} and Fig.~\ref{fig:Scaatter Map}, compared with reconstruction-based methods such as AE-Flow and DAE, the proposed model achieves adaptive reconstruction of regular features under the guidance of the INPs extractor, effectively suppressing incorrect reconstructions in anomalous regions and thereby improving both detection and localization performance. For instance, under the Flair modality (Combination 1), our method achieves image-level and pixel-level AUROC scores of 0.9472 and 0.9921, respectively, which represent significant improvements over AE-Flow (0.7359 and 0.8521) and DAE (0.8499 and 0.9598).

 For knowledge distillation-based models such as STFPM and RD4AD, the teacher model's generalization capacity strongly influences the student model's detection performance. In our framework, the DINOv2 encoder serves as a high-quality teacher, while the INPs-guided decoder reconstructs and aligns feature representations, forming a robust detection pipeline. Under the Flair modality, the proposed method improves image-level AP by 1.64 percentage  
\vspace{1em} 
\noindent
\includegraphics[width=\columnwidth]{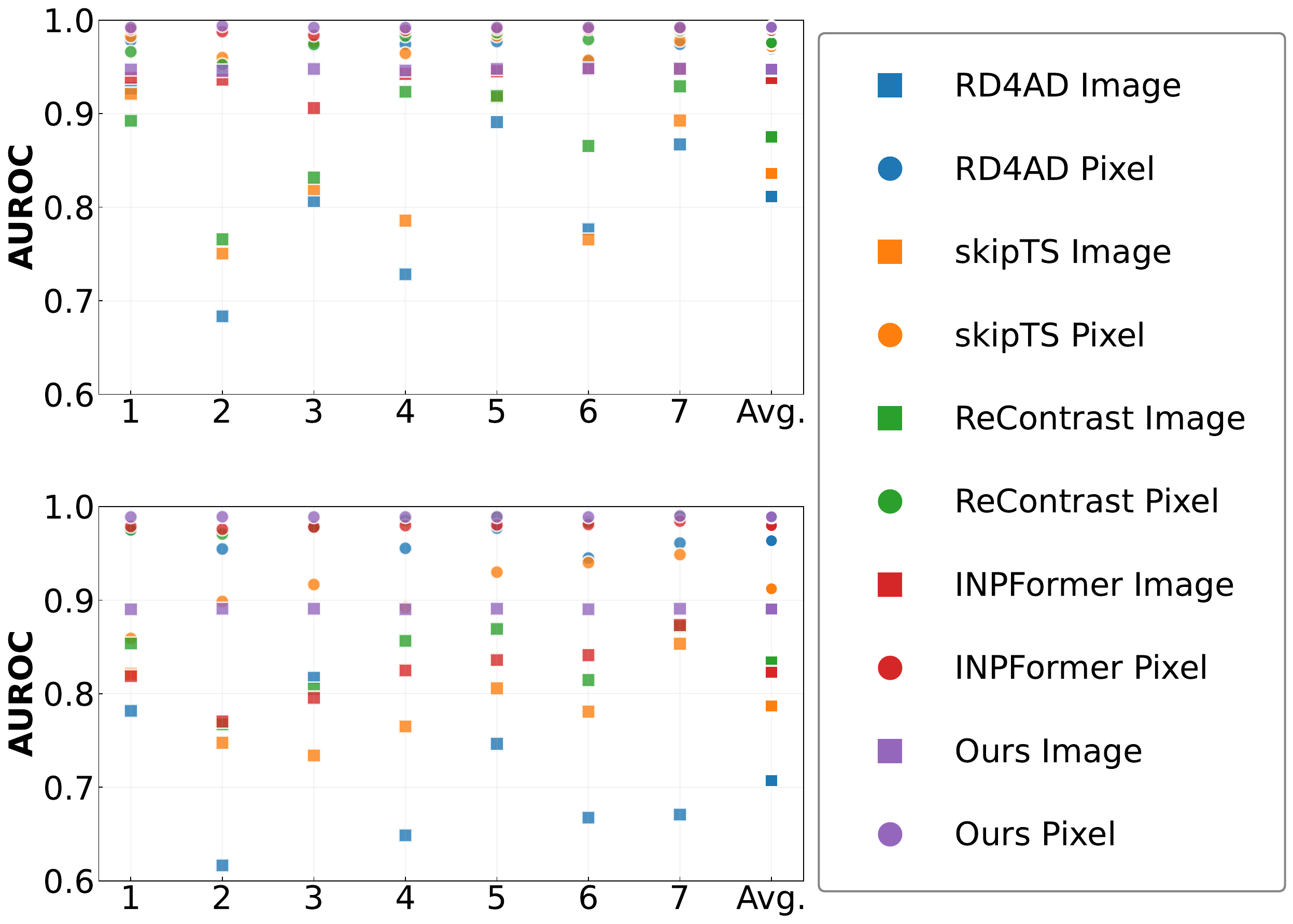}
\captionof{figure}{\normalfont\normalsize\textbf{Scatter plots comparing the AUROC performance of the top-5 models across the BraTS2018 (top row) and MU-Glioma-Post (bottom row) datasets}. Colors and shapes encode model identity and evaluation granularity, with the squares representing image-level and the circles representing pixel-level evaluation. The abscissa shows each modality combination 1-7 and the average performance, while the ordinate is the AUROC score.}
\vspace{1em} 
\label{fig:Scaatter Map}

\vspace{0.5em} 
\noindent
\begin{minipage}{\columnwidth} 
    \centering 
    \includegraphics[width=0.8\columnwidth]{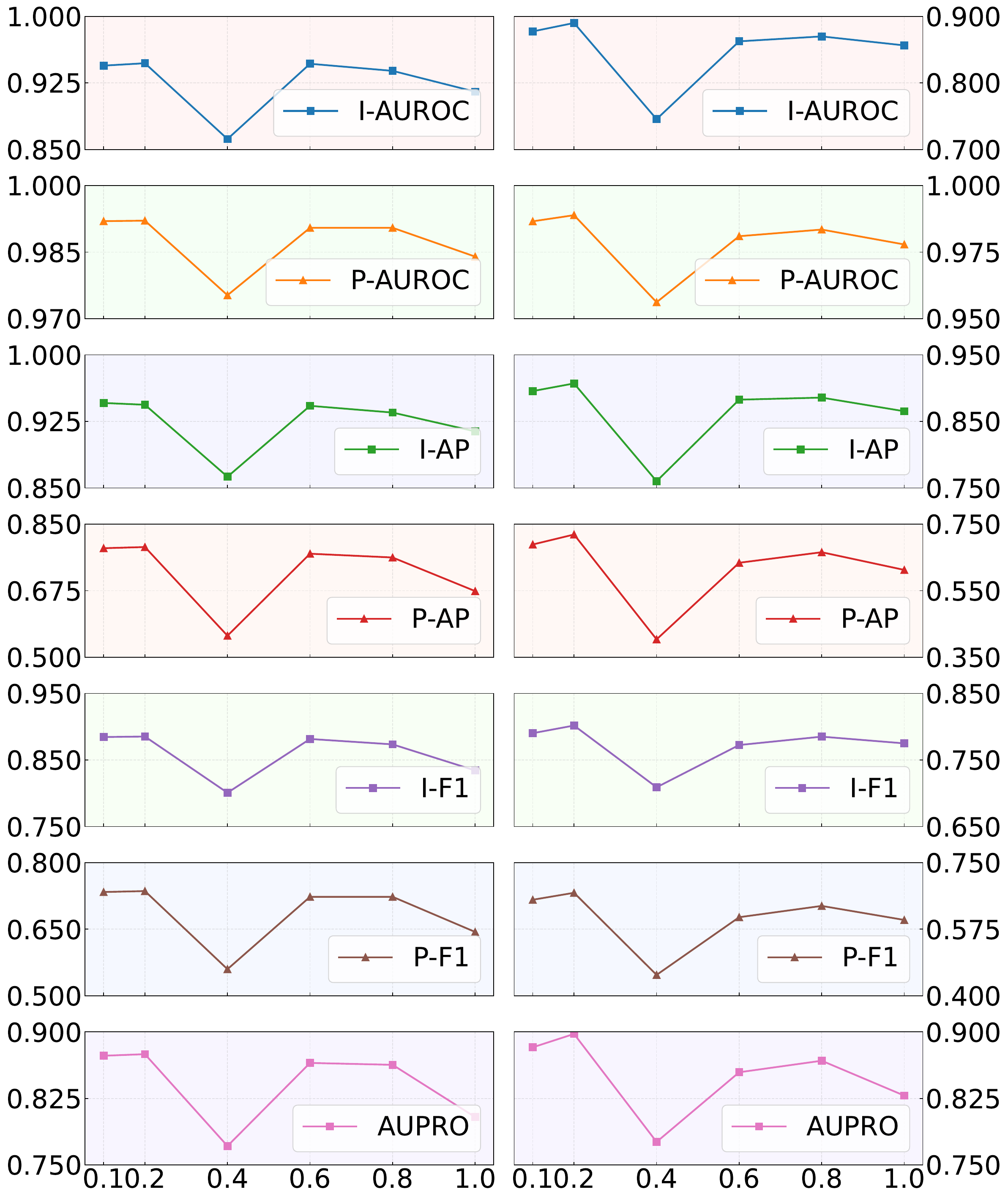}
    \captionof{figure}{\normalfont\normalsize\textbf{Analysis of the distribution alignment loss hy perparameter $\lambda_2$ on model performance across the BraTS2018 (left column) and MU-Glioma-Post (right column) datasets}. The left and right panels correspond to the respective datasets, each containing seven subplots of different performance metrics. All subplots are derived from the specific modality combination Combo1, with the x-axis indicating the precise values of $\lambda_2$ and the y-axis showing the corresponding quantitative performance scores.}
    \label{fig:line Map}
\end{minipage}
\vspace{1em}

points over RD4AD, increasing it from 0.9274 to 0.9438. This result demonstrates the synergistic effectiveness of prototype-guided reconstruction and distillation.
\hspace*{2em} Moreover, the proposed approach offers competitive advantages over the most recent state-of-the-art methods. When compared with the second-best model, ReContrast under the full-modality setting (Combination 7) improves the image-level AUROC from 0.9293 to 0.9482, the pixel-level AUROC from 0.9888 to 0.9921, and the AUPRO by 2.24 percentage points. At the same time, the visualization results in Fig. \ref{fig:BraTS Heat Map} and Fig. \ref{fig:MU Heat Map} also indicate that our method has excellent capability in anomaly localization. Relative to the baseline INP-Former, which already achieves strong performance under the Flair modality with image-level and pixel-level APs of 0.9369 and 0.7609, our model further increases these metrics to 0.9438 and 0.7898, respectively, corresponding to gains of 0.69 and 2.89 percentage points. The improvement primarily results from the introduction of the Feature Distribution Alignment mechanism, which enforces cross-modal feature consistency even in information-rich modalities, thereby enhancing detection robustness.

\hspace*{2em} It is worth noting that most existing models exhibit limited adaptability when facing missing-modality scenarios. To ensure a fair comparison, all competing methods were standardized to accept three-channel inputs, and the missing modalities were masked accordingly. However, these models generally require separate training for each modality combination, which results in non-shareable weights and limited scalability. Furthermore, their performance strongly depends on the richness of modality information. Although they perform well under high-information Flair inputs, their performance drops substantially when evaluated on T1 or T2 modalities. For example, the image-level AP of RD4AD decreases from 0.9274 under Flair to 0.6720 under T1, and INP-Former drops by 2.77 percentage points under T2.

\hspace*{2em} In contrast, our method supports unified weight learning across arbitrary modality combinations, significantly improving generalization and training efficiency. Benefiting from the proposed distribution alignment strategy, the model can automatically compensate for missing features in low-information modalities and maintain stable detection performance. Under the T2 modality, the image-level and pixel-level F1 scores increase from 0.8298 to 0.8841 and from 0.6391 to 0.7344, respectively, which validates the robustness of the proposed alignment mechanism. Consistent results obtained on the MU-Glioma-Post dataset further confirm the cross-dataset reliability and generalizability of the proposed approach, highlighting its practical value and research potential for in clinical applications.

\subsection{Ablation Experiments}
\hspace*{2em} To validate the effectiveness of the proposed distribution alignment loss function, a series of ablation experiments was conducted. While keeping the network architecture and training strategies unchanged, the distribution alignment loss was introduced separately to $En_0$ and $En_1$, and the results were compared with those obtained without this loss.

\hspace*{2em} As shown in Table \ref{table:ablation_2018} and Table \ref{table:ablationus_mu}, on both the BraTS2018 and MU-Glioma-Post datasets, the model consistently outperforms the baseline without the distribution alignment loss, regardless of whether the loss is applied to En0 or En1. When the loss is introduced at the En1 layer, the image-level and pixel-level APs under the Flair modality reach 0.9438 and 0.7898, respectively, representing clear improvements over the En0 layer (0.9366 and 0.7535) and the case without the loss (0.8528 and 0.5304). These results demonstrate that the distribution alignment loss effectively mitigates feature shift caused by missing modalities and, particularly at higher feature levels, enhances cross-modal consistency and improves the model’s generalization.

\begin{table*}[htbp]
\centering
\caption{\normalfont\normalsize Ablation experiments results on the BraTS2018 dataset.}
\vspace{0.5em}
\label{table:ablation_2018}
\resizebox{\textwidth}{!}{%
\begin{tabular}{@{}l l *{7}{c} c@{}}
\toprule
\textbf{Metric} & \textbf{Variant} &
\circletextgray{F}\circletext{T1}\circletext{T2} &
\circletext{F}\circletextgray{T1}\circletext{T2} &
\circletext{F}\circletext{T1}\circletextgray{T2} &
\circletextgray{F}\circletextgray{T1}\circletext{T2} &
\circletextgray{F}\circletext{T1}\circletextgray{T2} &
\circletext{F}\circletextgray{T1}\circletextgray{T2} &
\circletextgray{F}\circletextgray{T1}\circletextgray{T2} &
\textbf{Avg} \\
\midrule

\multirow{3}{*}{\textbf{AUROC}}
  & \cellcolor{lightyellow} En0           & \cellcolor{lightyellow!25} 0.9408/0.9895 & \cellcolor{lightyellow!25} 0.9408/0.9896 & \cellcolor{lightyellow!25} 0.9402/0.9895 & \cellcolor{lightyellow!25} 0.9407/0.9896 & \cellcolor{lightyellow!25} 0.9406/0.9895 & \cellcolor{lightyellow!25} 0.9403/0.9895 & \cellcolor{lightyellow!25} 0.9399/0.9895 & \cellcolor{lightyellow!25} 0.9405/0.9895 \\
  & \cellcolor{lightyellow} En1           & \cellcolor{lightyellow!25} 0.9472/0.9921 & \cellcolor{lightyellow!25} 0.9458/0.9939 & \cellcolor{lightyellow!25} 0.9479/0.9921 & \cellcolor{lightyellow!25} 0.9462/0.9920 & \cellcolor{lightyellow!25} 0.9478/0.9921 & \cellcolor{lightyellow!25} 0.9482/0.9921 & \cellcolor{lightyellow!25} 0.9481/0.9921 & \cellcolor{lightyellow!25} 0.9469/0.9923 \\
  & \cellcolor{lightyellow} w/o Loss & \cellcolor{lightyellow!25} 0.8529/0.9735 & \cellcolor{lightyellow!25} 0.8606/0.9739 & \cellcolor{lightyellow!25} 0.8495/0.9732 & \cellcolor{lightyellow!25} 0.8559/0.9737 & \cellcolor{lightyellow!25} 0.8511/0.9732 & \cellcolor{lightyellow!25} 0.8567/0.9740 & \cellcolor{lightyellow!25} 0.8583/0.9727 & \cellcolor{lightyellow!25} 0.8550/0.9735 \\
\hdashline

\multirow{3}{*}{\textbf{AP}}
  & \cellcolor{lightred} En0           & \cellcolor{lightred!25} 0.9366/0.7535 & \cellcolor{lightred!25} 0.9370/0.7564 & \cellcolor{lightred!25} 0.9360/0.7499 & \cellcolor{lightred!25} 0.9367/0.7544 & \cellcolor{lightred!25} 0.9363/0.7506 & \cellcolor{lightred!25} 0.9358/0.7492 & \cellcolor{lightred!25} 0.9356/0.7478 & \cellcolor{lightred!25} 0.9363/0.7517 \\
  & \cellcolor{lightred} En1           & \cellcolor{lightred!25} 0.9438/0.7898 & \cellcolor{lightred!25} 0.9429/0.7872 & \cellcolor{lightred!25} 0.9470/0.7880 & \cellcolor{lightred!25} 0.9433/0.7886 & \cellcolor{lightred!25} 0.9469/0.7886 & \cellcolor{lightred!25} 0.9472/0.7883 & \cellcolor{lightred!25} 0.9468/0.7860 & \cellcolor{lightred!25} 0.9454/0.7881 \\
  & \cellcolor{lightred} w/o Loss & \cellcolor{lightred!25} 0.8528/0.5304 & \cellcolor{lightred!25} 0.8575/0.5385 & \cellcolor{lightred!25} 0.8501/0.5251 & \cellcolor{lightred!25} 0.8543/0.5344 & \cellcolor{lightred!25} 0.8513/0.5262 & \cellcolor{lightred!25} 0.8556/0.5211 & \cellcolor{lightred!25} 0.8586/0.5386 & \cellcolor{lightred!25} 0.8543/0.5306 \\
\hdashline

\multirow{3}{*}{\textbf{F1}}
  & \cellcolor{lightpurple} En0           & \cellcolor{lightpurple!25} 0.8716/0.7073 & \cellcolor{lightpurple!25} 0.8767/0.7100 & \cellcolor{lightpurple!25} 0.8688/0.7044 & \cellcolor{lightpurple!25} 0.8743/0.7087 & \cellcolor{lightpurple!25} 0.8702/0.7051 & \cellcolor{lightpurple!25} 0.8711/0.7042 & \cellcolor{lightpurple!25} 0.8722/0.7024 & \cellcolor{lightpurple!25} 0.8721/0.7060 \\
  & \cellcolor{lightpurple} En1           & \cellcolor{lightpurple!25} 0.8852/0.7359 & \cellcolor{lightpurple!25} 0.8869/0.7344 & \cellcolor{lightpurple!25} 0.8881/0.7349 & \cellcolor{lightpurple!25} 0.8856/0.7352 & \cellcolor{lightpurple!25} 0.8877/0.7353 & \cellcolor{lightpurple!25} 0.8881/0.7349 & \cellcolor{lightpurple!25} 0.8849/0.7330 & \cellcolor{lightpurple!25} 0.8866/0.7348 \\
  & \cellcolor{lightpurple} w/o Loss & \cellcolor{lightpurple!25} 0.7829/0.5361 & \cellcolor{lightpurple!25} 0.7930/0.5430 & \cellcolor{lightpurple!25} 0.7822/0.5329 & \cellcolor{lightpurple!25} 0.7897/0.5396 & \cellcolor{lightpurple!25} 0.7844/0.5333 & \cellcolor{lightpurple!25} 0.7833/0.5307 & \cellcolor{lightpurple!25} 0.7951/0.5388 & \cellcolor{lightpurple!25} 0.7872/0.5363 \\
\hdashline

\multirow{3}{*}{\textbf{AUPRO}}
  & \cellcolor{darkblue} En0           & \cellcolor{darkblue!25} 0.8579 & \cellcolor{darkblue!25} 0.8608 & \cellcolor{darkblue!25} 0.8580 & \cellcolor{darkblue!25} 0.8589 & \cellcolor{darkblue!25} 0.8580 & \cellcolor{darkblue!25} 0.8586 & \cellcolor{darkblue!25} 0.8577 & \cellcolor{darkblue!25} 0.8584 \\
  & \cellcolor{darkblue} En1           & \cellcolor{darkblue!25} 0.8749 & \cellcolor{darkblue!25} 0.8768 & \cellcolor{darkblue!25} 0.8748 & \cellcolor{darkblue!25} 0.8757 & \cellcolor{darkblue!25} 0.8746 & \cellcolor{darkblue!25} 0.8752 & \cellcolor{darkblue!25} 0.8746 & \cellcolor{darkblue!25} 0.8752 \\
  & \cellcolor{darkblue} w/o Loss & \cellcolor{darkblue!25} 0.7606 & \cellcolor{darkblue!25} 0.7640 & \cellcolor{darkblue!25} 0.7585 & \cellcolor{darkblue!25} 0.7631 & \cellcolor{darkblue!25} 0.7581 & \cellcolor{darkblue!25} 0.7568 & \cellcolor{darkblue!25} 0.7637 & \cellcolor{darkblue!25} 0.7608 \\

\bottomrule
\end{tabular}%
}
\end{table*}

\begin{table*}[H]
\centering
\caption{\normalfont\normalsize Ablation experiments results on the MU-Glioma-Post dataset.}
\vspace{0.5em}
\label{table:ablationus_mu}
\resizebox{\textwidth}{!}{%
\begin{tabular}{@{}l l *{7}{c} c@{}}
\toprule
\textbf{Metric} & \textbf{Variant} &
\circletextgray{F}\circletext{T1}\circletext{T2} &
\circletext{F}\circletextgray{T1}\circletext{T2} &
\circletext{F}\circletext{T1}\circletextgray{T2} &
\circletextgray{F}\circletextgray{T1}\circletext{T2} &
\circletextgray{F}\circletext{T1}\circletextgray{T2} &
\circletext{F}\circletextgray{T1}\circletextgray{T2} &
\circletextgray{F}\circletextgray{T1}\circletextgray{T2} &
\textbf{Avg} \\
\midrule

\multirow{3}{*}{\textbf{AUROC}}
  & \cellcolor{lightyellow} En0           & \cellcolor{lightyellow!25} 0.8830/0.9852 & \cellcolor{lightyellow!25} 0.8835/0.9857 & \cellcolor{lightyellow!25} 0.8826/0.9850 & \cellcolor{lightyellow!25} 0.8835/0.9854 & \cellcolor{lightyellow!25} 0.8834/0.9851 & \cellcolor{lightyellow!25} 0.8816/0.9850 & \cellcolor{lightyellow!25} 0.8815/0.9849 & \cellcolor{lightyellow!25} 0.8827/0.9852 \\
  & \cellcolor{lightyellow} En1           & \cellcolor{lightyellow!25} 0.8901/0.9889 & \cellcolor{lightyellow!25} 0.8908/0.9891 & \cellcolor{lightyellow!25} 0.8909/0.9889 & \cellcolor{lightyellow!25} 0.8902/0.9889 & \cellcolor{lightyellow!25} 0.8907/0.9890 & \cellcolor{lightyellow!25} 0.8902/0.9889 & \cellcolor{lightyellow!25} 0.8908/0.9899 & \cellcolor{lightyellow!25} 0.8905/0.9891 \\
  & \cellcolor{lightyellow} w/o Loss & \cellcolor{lightyellow!25} 0.8292/0.9754 & \cellcolor{lightyellow!25} 0.8297/0.9765 & \cellcolor{lightyellow!25} 0.8292/0.9749 & \cellcolor{lightyellow!25} 0.8277/0.9758 & \cellcolor{lightyellow!25} 0.8297/0.9750 & \cellcolor{lightyellow!25} 0.8294/0.9746 & \cellcolor{lightyellow!25} 0.8292/0.9739 & \cellcolor{lightyellow!25} 0.8291/0.9751 \\
\hdashline

\multirow{3}{*}{\textbf{AP}}
  & \cellcolor{lightred} En0           & \cellcolor{lightred!25} 0.9015/0.6925 & \cellcolor{lightred!25} 0.9016/0.6950 & \cellcolor{lightred!25} 0.9010/0.6906 & \cellcolor{lightred!25} 0.9015/0.6930 & \cellcolor{lightred!25} 0.9000/0.6881 & \cellcolor{lightred!25} 0.9002/0.6884 & \cellcolor{lightred!25} 0.9018/0.6918 & \cellcolor{lightred!25} 0.9011/0.6913 \\
  & \cellcolor{lightred} En1           & \cellcolor{lightred!25} 0.9071/0.7188 & \cellcolor{lightred!25} 0.9071/0.7186 & \cellcolor{lightred!25} 0.9075/0.7195 & \cellcolor{lightred!25} 0.9069/0.7180 & \cellcolor{lightred!25} 0.9075/0.7199 & \cellcolor{lightred!25} 0.9067/0.7180 & \cellcolor{lightred!25} 0.9075/0.7201 & \cellcolor{lightred!25} 0.9072/0.7189 \\
  & \cellcolor{lightred} w/o Loss & \cellcolor{lightred!25} 0.8493/0.5879 & \cellcolor{lightred!25} 0.8509/0.5963 & \cellcolor{lightred!25} 0.8491/0.5813 & \cellcolor{lightred!25} 0.8484/0.5915 & \cellcolor{lightred!25} 0.8500/0.5831 & \cellcolor{lightred!25} 0.8484/0.5760 & \cellcolor{lightred!25} 0.8469/0.5693 & \cellcolor{lightred!25} 0.8490/0.5836 \\
\hdashline

\multirow{3}{*}{\textbf{F1}}
  & \cellcolor{lightpurple} En0           & \cellcolor{lightpurple!25} 0.7957/0.6531 & \cellcolor{lightpurple!25} 0.7947/0.6554 & \cellcolor{lightpurple!25} 0.7962/0.6515 & \cellcolor{lightpurple!25} 0.7942/0.6538 & \cellcolor{lightpurple!25} 0.7958/0.6524 & \cellcolor{lightpurple!25} 0.7952/0.6505 & \cellcolor{lightpurple!25} 0.7926/0.6503 & \cellcolor{lightpurple!25} 0.7949/0.6524 \\
  & \cellcolor{lightpurple} En1           & \cellcolor{lightpurple!25} 0.8017/0.6707 & \cellcolor{lightpurple!25} 0.8023/0.6701 & \cellcolor{lightpurple!25} 0.8021/0.6715 & \cellcolor{lightpurple!25} 0.8017/0.6698 & \cellcolor{lightpurple!25} 0.8012/0.6713 & \cellcolor{lightpurple!25} 0.8017/0.6708 & \cellcolor{lightpurple!25} 0.8035/0.6726 & \cellcolor{lightpurple!25} 0.8020/0.6710 \\
  & \cellcolor{lightpurple} w/o Loss & \cellcolor{lightpurple!25} 0.7610/0.5770 & \cellcolor{lightpurple!25} 0.7581/0.5828 & \cellcolor{lightpurple!25} 0.7593/0.5725 & \cellcolor{lightpurple!25} 0.7605/0.5791 & \cellcolor{lightpurple!25} 0.7606/0.5736 & \cellcolor{lightpurple!25} 0.7584/0.5694 & \cellcolor{lightpurple!25} 0.7566/0.5651 & \cellcolor{lightpurple!25} 0.7593/0.5742 \\
\hdashline

\multirow{3}{*}{\textbf{AUPRO}}
  & \cellcolor{darkblue} En0           & \cellcolor{darkblue!25} 0.8760 & \cellcolor{darkblue!25} 0.8795 & \cellcolor{darkblue!25} 0.8748 & \cellcolor{darkblue!25} 0.8773 & \cellcolor{darkblue!25} 0.8756 & \cellcolor{darkblue!25} 0.8738 & \cellcolor{darkblue!25} 0.8733 & \cellcolor{darkblue!25} 0.8758 \\
  & \cellcolor{darkblue} En1           & \cellcolor{darkblue!25} 0.8977 & \cellcolor{darkblue!25} 0.9007 & \cellcolor{darkblue!25} 0.8973 & \cellcolor{darkblue!25} 0.8987 & \cellcolor{darkblue!25} 0.8978 & \cellcolor{darkblue!25} 0.8967 & \cellcolor{darkblue!25} 0.9012 & \cellcolor{darkblue!25} 0.8986 \\
  & \cellcolor{darkblue} w/o Loss & \cellcolor{darkblue!25} 0.8124 & \cellcolor{darkblue!25} 0.8199 & \cellcolor{darkblue!25} 0.8091 & \cellcolor{darkblue!25} 0.8152 & \cellcolor{darkblue!25} 0.8091 & \cellcolor{darkblue!25} 0.8067 & \cellcolor{darkblue!25} 0.8035 & \cellcolor{darkblue!25} 0.8108 \\

\bottomrule
\end{tabular}%
}
\end{table*}

\hspace*{2em} Furthermore, the influence of the loss weight hyperparameter $\lambda_2$ was analyzed in detail, as shown in Table~\ref{table:lossval_2018} and Table~\ref{table:lossval_mu}. The line plot in Fig.~\ref{fig:line Map} illustrates the variation in model performance under different weight configurations. When $\lambda_2 = 0.2$, the model achieves optimal performance on both the BraTS2018 and MU-Glioma-Post datasets, with image-level and pixel-level APs of 0.9438 and 0.7898, respectively, under the Flair modality. When the weight increases excessively, for example, when $\lambda_2 = 0.4$, the model performance decreases significantly to 0.8627 and 0.5566. This observation indicates that appropriately balancing the distribution alignment loss with the main task loss is essential for maintaining performance stability. Based on these experimental findings, $\lambda_2$ is set to 0.2 in this study.

\begin{table*}[htbp]
\centering
\caption{\normalfont\normalsize Model performance of the hyperparameter $\lambda_2$ on the BraTS2018 dataset.}
\vspace{0.5em}
\label{table:lossval_2018}
\resizebox{\textwidth}{!}{%
\begin{tabular}{@{}l l *{7}{c} c@{}}
\toprule
\textbf{Metrics} & $\lambda_2$\ & \circletextgray{F}\circletext{T1}\circletext{T2} & \circletext{F}\circletextgray{T1}\circletext{T2} & \circletext{F}\circletext{T1}\circletextgray{T2} & \circletextgray{F}\circletextgray{T1}\circletext{T2} & \circletextgray{F}\circletext{T1}\circletextgray{T2} & \circletext{F}\circletextgray{T1}\circletextgray{T2} & \circletextgray{F}\circletextgray{T1}\circletextgray{T2} & \textbf{Avg} \\
\midrule

\multirow{6}{*}{\textbf{AUROC}} 
  & \cellcolor{lightyellow} 0.1 & \cellcolor{lightyellow!25} 0.9445/0.9920 & \cellcolor{lightyellow!25} 0.9442/0.9919 & \cellcolor{lightyellow!25} 0.9446/0.9920 & \cellcolor{lightyellow!25} 0.9443/0.9920 & \cellcolor{lightyellow!25} 0.9448/0.9920 & \cellcolor{lightyellow!25} 0.9445/0.9920 & \cellcolor{lightyellow!25} 0.9448/0.9920 & \cellcolor{lightyellow!25} 0.9445/0.9920 \\
  & \cellcolor{lightyellow}\ 0.2 & \cellcolor{lightyellow!25}\color{red} 0.9472/0.9921 & \cellcolor{lightyellow!25}\color{red} 0.9458/0.9939 & \cellcolor{lightyellow!25}\color{red} 0.9479/0.9921 & \cellcolor{lightyellow!25}\color{red} 0.9462/0.9920 & \cellcolor{lightyellow!25}\color{red} 0.9478/0.9921 & \cellcolor{lightyellow!25}\color{red} 0.9482/0.9921 & \cellcolor{lightyellow!25}\color{red} 0.9481/0.9921 & \cellcolor{lightyellow!25}\color{red} 0.9469/0.9923 \\
  & \cellcolor{lightyellow} 0.4 & \cellcolor{lightyellow!25} 0.8618/0.9753 & \cellcolor{lightyellow!25} 0.8650/0.9768 & \cellcolor{lightyellow!25} 0.8613/0.9747 & \cellcolor{lightyellow!25} 0.8613/0.9758 & \cellcolor{lightyellow!25} 0.8615/0.9747 & \cellcolor{lightyellow!25} 0.8589/0.9743 & \cellcolor{lightyellow!25} 0.8618/0.9747 & \cellcolor{lightyellow!25} 0.8617/0.9752 \\
  & \cellcolor{lightyellow} 0.6 & \cellcolor{lightyellow!25} 0.9466/0.9905 & \cellcolor{lightyellow!25} 0.9447/0.9903 & \cellcolor{lightyellow!25} 0.9458/0.9904 & \cellcolor{lightyellow!25} 0.9457/0.9904 & \cellcolor{lightyellow!25} 0.9464/0.9904 & \cellcolor{lightyellow!25} 0.9468/0.9904 & \cellcolor{lightyellow!25} 0.9471/0.9904 & \cellcolor{lightyellow!25} 0.9387/0.9905 \\
  & \cellcolor{lightyellow} 0.8 & \cellcolor{lightyellow!25} 0.9386/0.9905 & \cellcolor{lightyellow!25} 0.9374/0.9904 & \cellcolor{lightyellow!25} 0.9392/0.9904 & \cellcolor{lightyellow!25} 0.9376/0.9904 & \cellcolor{lightyellow!25} 0.9393/0.9905 & \cellcolor{lightyellow!25} 0.9391/0.9905 & \cellcolor{lightyellow!25} 0.9416/0.9907 & \cellcolor{lightyellow!25} 0.9387/0.9905 \\
  & \cellcolor{lightyellow} 1.0 & \cellcolor{lightyellow!25} 0.9151/0.9840 & \cellcolor{lightyellow!25} 0.9155/0.9845 & \cellcolor{lightyellow!25} 0.9152/0.9836 & \cellcolor{lightyellow!25} 0.9150/0.9843 & \cellcolor{lightyellow!25} 0.9157/0.9837 & \cellcolor{lightyellow!25} 0.9157/0.9836 & \cellcolor{lightyellow!25} 0.9162/0.9834 & \cellcolor{lightyellow!25} 0.9155/0.9839 \\
\hdashline
\multirow{6}{*}{\textbf{AP}}
  & \cellcolor{lightred} 0.1 & \cellcolor{lightred!25} 0.9458/0.7867 & \cellcolor{lightred!25} 0.9438/0.7867 & \cellcolor{lightred!25} 0.9458/0.7849 & \cellcolor{lightred!25} 0.9457/0.7876 & \cellcolor{lightred!25} 0.9457/0.7858 & \cellcolor{lightred!25} 0.9456/0.7850 & \cellcolor{lightred!25} 0.9460/0.7867 & \cellcolor{lightred!25} 0.9455/0.7862 \\
  & \cellcolor{lightred}\ 0.2 & \cellcolor{lightred!25}\color{red} 0.9438/0.7898 & \cellcolor{lightred!25}\color{red} 0.9429/0.7872 & \cellcolor{lightred!25}\color{red} 0.9470/0.7880 & \cellcolor{lightred!25}\color{red} 0.9433/0.7886 & \cellcolor{lightred!25}\color{red} 0.9469/0.7886 & \cellcolor{lightred!25}\color{red} 0.9472/0.7883 & \cellcolor{lightred!25}\color{red} 0.9468/0.7860 & \cellcolor{lightred!25}\color{red} 0.9454/0.7881 \\
  & \cellcolor{lightred} 0.4 & \cellcolor{lightred!25} 0.8627/0.5566 & \cellcolor{lightred!25} 0.8616/0.5665 & \cellcolor{lightred!25} 0.8621/0.5501 & \cellcolor{lightred!25} 0.8604/0.5600 & \cellcolor{lightred!25} 0.8625/0.5499 & \cellcolor{lightred!25} 0.8594/0.5445 & \cellcolor{lightred!25} 0.8640/0.5498 & \cellcolor{lightred!25} 0.8618/0.5539 \\
  & \cellcolor{lightred} 0.6 & \cellcolor{lightred!25} 0.9426/0.7722 & \cellcolor{lightred!25} 0.9398/0.7744 & \cellcolor{lightred!25} 0.9418/0.7760 & \cellcolor{lightred!25} 0.9402/0.7776 & \cellcolor{lightred!25} 0.9424/0.7750 & \cellcolor{lightred!25} 0.9425/0.7749 & \cellcolor{lightred!25} 0.9425/0.7738 & \cellcolor{lightred!25} 0.9417/0.7748 \\
  & \cellcolor{lightred} 0.8 & \cellcolor{lightred!25} 0.9350/0.7623 & \cellcolor{lightred!25} 0.9342/0.7626 & \cellcolor{lightred!25} 0.9352/0.7590 & \cellcolor{lightred!25} 0.9342/0.7631 & \cellcolor{lightred!25} 0.9353/0.7612 & \cellcolor{lightred!25} 0.9351/0.7585 & \cellcolor{lightred!25} 0.9365/0.7608 & \cellcolor{lightred!25} 0.9351/0.7611 \\
  & \cellcolor{lightred} 1.0 & \cellcolor{lightred!25} 0.9138/0.6739 & \cellcolor{lightred!25} 0.9149/0.6815 & \cellcolor{lightred!25} 0.9138/0.6674 & \cellcolor{lightred!25} 0.9143/0.6792 & \cellcolor{lightred!25} 0.9140/0.6701 & \cellcolor{lightred!25} 0.9140/0.6672 & \cellcolor{lightred!25} 0.9140/0.6634 & \cellcolor{lightred!25} 0.9141/0.6718 \\
\hdashline

\multirow{6}{*}{\textbf{F1}}
  & \cellcolor{lightpurple} 0.1 & \cellcolor{lightpurple!25} 0.8844/0.7338 & \cellcolor{lightpurple!25} 0.8869/0.7326 & \cellcolor{lightpurple!25} 0.8832/0.7323 & \cellcolor{lightpurple!25} 0.8840/0.7339 & \cellcolor{lightpurple!25} 0.8841/0.7329 & \cellcolor{lightpurple!25} 0.8848/0.7323 & \cellcolor{lightpurple!25} 0.8852/0.7339 & \cellcolor{lightpurple!25} 0.8847/0.7331 \\
  & \cellcolor{lightpurple}\ 0.2 & \cellcolor{lightpurple!25}\color{red} 0.8852/0.7359 & \cellcolor{lightpurple!25}\color{red} 0.8869/0.7344 & \cellcolor{lightpurple!25}\color{red} 0.8881/0.7349 & \cellcolor{lightpurple!25}\color{red} 0.8856/0.7352 & \cellcolor{lightpurple!25}\color{red} 0.8877/0.7353 & \cellcolor{lightpurple!25}\color{red} 0.8881/0.7349 & \cellcolor{lightpurple!25}\color{red} 0.8849/0.7330 & \cellcolor{lightpurple!25}\color{red} 0.7981/0.5586 \\
  & \cellcolor{lightpurple} 0.4 & \cellcolor{lightpurple!25} 0.8007/0.5599 & \cellcolor{lightpurple!25} 0.8022/0.5686 & \cellcolor{lightpurple!25} 0.7986/0.5558 & \cellcolor{lightpurple!25} 0.7993/0.5628 & \cellcolor{lightpurple!25} 0.7973/0.5555 & \cellcolor{lightpurple!25} 0.7922/0.5527 & \cellcolor{lightpurple!25} 0.7966/0.5552 & \cellcolor{lightpurple!25} 0.7981/0.5586 \\
  & \cellcolor{lightpurple} 0.6 & \cellcolor{lightpurple!25} 0.8815/0.7230 & \cellcolor{lightpurple!25} 0.8841/0.7202 & \cellcolor{lightpurple!25} 0.8815/0.7217 & \cellcolor{lightpurple!25} 0.8824/0.7231 & \cellcolor{lightpurple!25} 0.8822/0.7205 & \cellcolor{lightpurple!25} 0.8807/0.7209 & \cellcolor{lightpurple!25} 0.8826/0.7200 & \cellcolor{lightpurple!25} 0.9351/0.7611 \\
  & \cellcolor{lightpurple} 0.8 & \cellcolor{lightpurple!25} 0.8734/0.7230 & \cellcolor{lightpurple!25} 0.8768/0.7233 & \cellcolor{lightpurple!25} 0.8766/0.7210 & \cellcolor{lightpurple!25} 0.8784/0.7242 & \cellcolor{lightpurple!25} 0.8741/0.7223 & \cellcolor{lightpurple!25} 0.8750/0.7202 & \cellcolor{lightpurple!25} 0.8739/0.7216 & \cellcolor{lightpurple!25} 0.8754/0.7222 \\
  & \cellcolor{lightpurple} 1.0 & \cellcolor{lightpurple!25} 0.8342/0.6438 & \cellcolor{lightpurple!25} 0.8406/0.6497 & \cellcolor{lightpurple!25} 0.8325/0.6388 & \cellcolor{lightpurple!25} 0.8383/0.6479 & \cellcolor{lightpurple!25} 0.8345/0.6406 & \cellcolor{lightpurple!25} 0.8328/0.6382 & \cellcolor{lightpurple!25} 0.8308/0.6354 & \cellcolor{lightpurple!25} 0.8348/0.6420 \\
\hdashline

\multirow{6}{*}{\textbf{AUPRO}}
  & \cellcolor{darkblue} 0.1 & \cellcolor{darkblue!25} 0.8731 & \cellcolor{darkblue!25} 0.8747 & \cellcolor{darkblue!25} 0.8733 & \cellcolor{darkblue!25} 0.8737 & \cellcolor{darkblue!25} 0.8729 & \cellcolor{darkblue!25} 0.8734 & \cellcolor{darkblue!25} 0.8747 & \cellcolor{darkblue!25} 0.8734 \\
  & \cellcolor{darkblue}\ 0.2 & \cellcolor{darkblue!25}\color{red} 0.8749 & \cellcolor{darkblue!25}\color{red} 0.8768 & \cellcolor{darkblue!25}\color{red} 0.8748 & \cellcolor{darkblue!25}\color{red} 0.8757 & \cellcolor{darkblue!25}\color{red} 0.8746 & \cellcolor{darkblue!25}\color{red} 0.8752 & \cellcolor{darkblue!25}\color{red} 0.8746 & \cellcolor{darkblue!25}\color{red} 0.8752 \\
  & \cellcolor{darkblue} 0.4 & \cellcolor{darkblue!25} 0.7714 & \cellcolor{darkblue!25} 0.7714 & \cellcolor{darkblue!25} 0.7714 & \cellcolor{darkblue!25} 0.7714 & \cellcolor{darkblue!25} 0.7621 & \cellcolor{darkblue!25} 0.7605 & \cellcolor{darkblue!25} 0.7631 & \cellcolor{darkblue!25} 0.7673 \\
  & \cellcolor{darkblue} 0.6 & \cellcolor{darkblue!25} 0.8650 & \cellcolor{darkblue!25} 0.8666 & \cellcolor{darkblue!25} 0.8637 & \cellcolor{darkblue!25} 0.8658 & \cellcolor{darkblue!25} 0.8636 & \cellcolor{darkblue!25} 0.8639 & \cellcolor{darkblue!25} 0.8641 & \cellcolor{darkblue!25} 0.8647 \\
  & \cellcolor{darkblue} 0.8 & \cellcolor{darkblue!25} 0.8629 & \cellcolor{darkblue!25} 0.8638 & \cellcolor{darkblue!25} 0.8638 & \cellcolor{darkblue!25} 0.8622 & \cellcolor{darkblue!25} 0.8629 & \cellcolor{darkblue!25} 0.8631 & \cellcolor{darkblue!25} 0.8642 & \cellcolor{darkblue!25} 0.8633 \\
  & \cellcolor{darkblue} 1.0 & \cellcolor{darkblue!25} 0.8040 & \cellcolor{darkblue!25} 0.8100 & \cellcolor{darkblue!25} 0.8100 & \cellcolor{darkblue!25} 0.8160 & \cellcolor{darkblue!25} 0.8112 & \cellcolor{darkblue!25} 0.8111 & \cellcolor{darkblue!25} 0.8089 & \cellcolor{darkblue!25} 0.8087 \\
\bottomrule
\end{tabular}%
}
\end{table*}

\begin{table*}[H]
\centering
\caption{\normalfont\normalsize Model performance of the hyperparameter $\lambda_2$ on the MU-Glioma-Post dataset.}
\vspace{0.5em}
\label{table:lossval_mu}
\resizebox{\textwidth}{!}{%
\begin{tabular}{@{}l l *{7}{c} c@{}}
\toprule
\textbf{Metrics} & $\lambda_2$ & 
\circletextgray{F}\circletext{T1}\circletext{T2} & 
\circletext{F}\circletextgray{T1}\circletext{T2} & 
\circletext{F}\circletext{T1}\circletextgray{T2} & 
\circletextgray{F}\circletextgray{T1}\circletext{T2} & 
\circletextgray{F}\circletext{T1}\circletextgray{T2} & 
\circletext{F}\circletextgray{T1}\circletextgray{T2} & 
\circletextgray{F}\circletextgray{T1}\circletextgray{T2} & 
\textbf{Avg} \\
\midrule

\multirow{6}{*}{\textbf{AUROC}} 
  & \cellcolor{lightyellow} 0.1 & \cellcolor{lightyellow!25} 0.8774/0.9866 & \cellcolor{lightyellow!25} 0.8775/0.9869 & \cellcolor{lightyellow!25} 0.8772/0.9866 & \cellcolor{lightyellow!25} 0.8770/0.9867 & \cellcolor{lightyellow!25} 0.8774/0.9866 & \cellcolor{lightyellow!25} 0.8765/0.9866 & \cellcolor{lightyellow!25} 0.8773/0.9867 & \cellcolor{lightyellow!25} 0.8772/0.9867 \\
  & \cellcolor{lightyellow}\ 0.2 & \cellcolor{lightyellow!25}\color{red} 0.8901/0.9889 & \cellcolor{lightyellow!25}\color{red} 0.8908/0.9891 & \cellcolor{lightyellow!25}\color{red} 0.8909/0.9889 & \cellcolor{lightyellow!25}\color{red} 0.8902/0.9889 & \cellcolor{lightyellow!25}\color{red} 0.8907/0.9890 & \cellcolor{lightyellow!25}\color{red} 0.8902/0.9889 & \cellcolor{lightyellow!25}\color{red} 0.8908/0.9899 & \cellcolor{lightyellow!25}\color{red} 0.8905/0.9891 \\
  & \cellcolor{lightyellow} 0.4 & \cellcolor{lightyellow!25} 0.7456/0.9562 & \cellcolor{lightyellow!25} 0.7478/0.9598 & \cellcolor{lightyellow!25} 0.7427/0.9546 & \cellcolor{lightyellow!25} 0.7446/0.9576 & \cellcolor{lightyellow!25} 0.7413/0.9548 & \cellcolor{lightyellow!25} 0.7464/0.9534 & \cellcolor{lightyellow!25} 0.7375/0.9520 & \cellcolor{lightyellow!25} 0.7437/0.9556 \\
  & \cellcolor{lightyellow} 0.6 & \cellcolor{lightyellow!25} 0.8625/0.9810 & \cellcolor{lightyellow!25} 0.8623/0.9816 & \cellcolor{lightyellow!25} 0.8636/0.9807 & \cellcolor{lightyellow!25} 0.8597/0.9811 & \cellcolor{lightyellow!25} 0.8641/0.9808 & \cellcolor{lightyellow!25} 0.8633/0.9805 & \cellcolor{lightyellow!25} 0.8628/0.9802 & \cellcolor{lightyellow!25} 0.8626/0.9808 \\
  & \cellcolor{lightyellow} 0.8 & \cellcolor{lightyellow!25} 0.8699/0.9835 & \cellcolor{lightyellow!25} 0.8688/0.9837 & \cellcolor{lightyellow!25} 0.8692/0.9835 & \cellcolor{lightyellow!25} 0.8687/0.9835 & \cellcolor{lightyellow!25} 0.8696/0.9837 & \cellcolor{lightyellow!25} 0.8695/0.9834 & \cellcolor{lightyellow!25} 0.8696/0.9835 & \cellcolor{lightyellow!25} 0.8693/0.9835 \\
  & \cellcolor{lightyellow} 1.0 & \cellcolor{lightyellow!25} 0.8564/0.9779 & \cellcolor{lightyellow!25} 0.8450/0.9778 & \cellcolor{lightyellow!25} 0.8506/0.9779 & \cellcolor{lightyellow!25} 0.8443/0.9774 & \cellcolor{lightyellow!25} 0.8514/0.9781 & \cellcolor{lightyellow!25} 0.8521/0.9779 & \cellcolor{lightyellow!25} 0.8546/0.9785 & \cellcolor{lightyellow!25} 0.8506/0.9778 \\
\hdashline

\multirow{6}{*}{\textbf{AP}}
  & \cellcolor{lightred} 0.1 & \cellcolor{lightred!25} 0.8956/0.6888 & \cellcolor{lightred!25} 0.8957/0.6903 & \cellcolor{lightred!25} 0.8951/0.6873 & \cellcolor{lightred!25} 0.8952/0.6889 & \cellcolor{lightred!25} 0.8956/0.6888 & \cellcolor{lightred!25} 0.8941/0.6854 & \cellcolor{lightred!25} 0.8949/0.6872 & \cellcolor{lightred!25} 0.8952/0.6879 \\
  & \cellcolor{lightred}\ 0.2 & \cellcolor{lightred!25}\color{red} 0.9071/0.7188 & \cellcolor{lightred!25}\color{red} 0.9071/0.7186 & \cellcolor{lightred!25}\color{red} 0.9075/0.7195 & \cellcolor{lightred!25}\color{red} 0.9069/0.7180 & \cellcolor{lightred!25}\color{red} 0.9075/0.7199 & \cellcolor{lightred!25}\color{red} 0.9067/0.7180 & \cellcolor{lightred!25}\color{red} 0.9075/0.7201 & \cellcolor{lightred!25}\color{red} 0.9072/0.7189 \\
  & \cellcolor{lightred} 0.4 & \cellcolor{lightred!25} 0.7602/0.4037 & \cellcolor{lightred!25} 0.7619/0.4132 & \cellcolor{lightred!25} 0.7563/0.3960 & \cellcolor{lightred!25} 0.7592/0.4075 & \cellcolor{lightred!25} 0.7555/0.3964 & \cellcolor{lightred!25} 0.7532/0.3889 & \cellcolor{lightred!25} 0.7493/0.3773 & \cellcolor{lightred!25} 0.7565/0.3976 \\
  & \cellcolor{lightred} 0.6 & \cellcolor{lightred!25} 0.8827/0.6339 & \cellcolor{lightred!25} 0.8830/0.6378 & \cellcolor{lightred!25} 0.8827/0.6310 & \cellcolor{lightred!25} 0.8806/0.6351 & \cellcolor{lightred!25} 0.8833/0.6319 & \cellcolor{lightred!25} 0.8819/0.6270 & \cellcolor{lightred!25} 0.8815/0.6244 & \cellcolor{lightred!25} 0.8822/0.6317 \\
  & \cellcolor{lightred} 0.8 & \cellcolor{lightred!25} 0.8858/0.6656 & \cellcolor{lightred!25} 0.8849/0.6637 & \cellcolor{lightred!25} 0.8850/0.6660 & \cellcolor{lightred!25} 0.8844/0.6637 & \cellcolor{lightred!25} 0.8859/0.6670 & \cellcolor{lightred!25} 0.8852/0.6633 & \cellcolor{lightred!25} 0.8848/0.6639 & \cellcolor{lightred!25} 0.8851/0.6647 \\
  & \cellcolor{lightred} 1.0 & \cellcolor{lightred!25} 0.8653/0.6124 & \cellcolor{lightred!25} 0.8612/0.6078 & \cellcolor{lightred!25} 0.8653/0.6121 & \cellcolor{lightred!25} 0.8605/0.6074 & \cellcolor{lightred!25} 0.8670/0.6156 & \cellcolor{lightred!25} 0.8650/0.6108 & \cellcolor{lightred!25} 0.8687/0.6201 & \cellcolor{lightred!25} 0.8647/0.6123 \\
\hdashline

\multirow{6}{*}{\textbf{F1}}
  & \cellcolor{lightpurple} 0.1 & \cellcolor{lightpurple!25} 0.7902/0.6524 & \cellcolor{lightpurple!25} 0.7873/0.6537 & \cellcolor{lightpurple!25} 0.7929/0.6517 & \cellcolor{lightpurple!25} 0.7895/0.6529 & \cellcolor{lightpurple!25} 0.7901/0.6525 & \cellcolor{lightpurple!25} 0.7929/0.6511 & \cellcolor{lightpurple!25} 0.7923/0.6520 & \cellcolor{lightpurple!25} 0.7912/0.6523 \\
  & \cellcolor{lightpurple}\ 0.2 & \cellcolor{lightpurple!25}\color{red} 0.8017/0.6707 & \cellcolor{lightpurple!25}\color{red} 0.8023/0.6701 & \cellcolor{lightpurple!25}\color{red} 0.8021/0.6715 & \cellcolor{lightpurple!25}\color{red} 0.8017/0.6698 & \cellcolor{lightpurple!25}\color{red} 0.8012/0.6713 & \cellcolor{lightpurple!25}\color{red} 0.8017/0.6708 & \cellcolor{lightpurple!25}\color{red} 0.8035/0.6726 & \cellcolor{lightpurple!25}\color{red} 0.8020/0.6710 \\
  & \cellcolor{lightpurple} 0.4 & \cellcolor{lightpurple!25} 0.7089/0.4548 & \cellcolor{lightpurple!25} 0.7102/0.4422 & \cellcolor{lightpurple!25} 0.7054/0.4296 & \cellcolor{lightpurple!25} 0.7115/0.4377 & \cellcolor{lightpurple!25} 0.7042/0.4295 & \cellcolor{lightpurple!25} 0.7020/0.4241 & \cellcolor{lightpurple!25} 0.6968/0.4164 & \cellcolor{lightpurple!25} 0.7056/0.4363 \\
  & \cellcolor{lightpurple} 0.6 & \cellcolor{lightpurple!25} 0.7724/0.6065 & \cellcolor{lightpurple!25} 0.7728/0.6098 & \cellcolor{lightpurple!25} 0.7742/0.6039 & \cellcolor{lightpurple!25} 0.7699/0.6080 & \cellcolor{lightpurple!25} 0.7757/0.6051 & \cellcolor{lightpurple!25} 0.7776/0.6013 & \cellcolor{lightpurple!25} 0.7760/0.5991 & \cellcolor{lightpurple!25} 0.7741/0.6048 \\
  & \cellcolor{lightpurple} 0.8 & \cellcolor{lightpurple!25} 0.7850/0.6362 & \cellcolor{lightpurple!25} 0.7849/0.6355 & \cellcolor{lightpurple!25} 0.7869/0.6361 & \cellcolor{lightpurple!25} 0.7863/0.6357 & \cellcolor{lightpurple!25} 0.7866/0.6371 & \cellcolor{lightpurple!25} 0.7880/0.6350 & \cellcolor{lightpurple!25} 0.7901/0.6358 & \cellcolor{lightpurple!25} 0.7868/0.6361 \\
  & \cellcolor{lightpurple} 1.0 & \cellcolor{lightpurple!25} 0.7749/0.5993 & \cellcolor{lightpurple!25} 0.7696/0.5900 & \cellcolor{lightpurple!25} 0.7784/0.5983 & \cellcolor{lightpurple!25} 0.7736/0.5956 & \cellcolor{lightpurple!25} 0.7840/0.6089 & \cellcolor{lightpurple!25} 0.7876/0.5978 & \cellcolor{lightpurple!25} 0.7820/0.6050 & \cellcolor{lightpurple!25} 0.7786/0.5993 \\
\hdashline

\multirow{6}{*}{\textbf{AUPRO}}
  & \cellcolor{darkblue} 0.1 & \cellcolor{darkblue!25} 0.8826 & \cellcolor{darkblue!25} 0.8851 & \cellcolor{darkblue!25} 0.8818 & \cellcolor{darkblue!25} 0.8835 & \cellcolor{darkblue!25} 0.8825 & \cellcolor{darkblue!25} 0.8814 & \cellcolor{darkblue!25} 0.8817 & \cellcolor{darkblue!25} 0.8827 \\
  & \cellcolor{darkblue}\ 0.2 & \cellcolor{darkblue!25}\color{red} 0.8977 & \cellcolor{darkblue!25}\color{red} 0.9007 & \cellcolor{darkblue!25}\color{red} 0.8973 & \cellcolor{darkblue!25}\color{red} 0.8987 & \cellcolor{darkblue!25}\color{red} 0.8978 & \cellcolor{darkblue!25}\color{red} 0.8967 & \cellcolor{darkblue!25}\color{red} 0.9012 & \cellcolor{darkblue!25}\color{red} 0.8986 \\
  & \cellcolor{darkblue} 0.4 & \cellcolor{darkblue!25} 0.7760 & \cellcolor{darkblue!25} 0.7911 & \cellcolor{darkblue!25} 0.7704 & \cellcolor{darkblue!25} 0.7820 & \cellcolor{darkblue!25} 0.7710 & \cellcolor{darkblue!25} 0.7660 & \cellcolor{darkblue!25} 0.7610 & \cellcolor{darkblue!25} 0.7739 \\
  & \cellcolor{darkblue} 0.6 & \cellcolor{darkblue!25} 0.8545 & \cellcolor{darkblue!25} 0.8582 & \cellcolor{darkblue!25} 0.8525 & \cellcolor{darkblue!25} 0.8552 & \cellcolor{darkblue!25} 0.8533 & \cellcolor{darkblue!25} 0.8515 & \cellcolor{darkblue!25} 0.8406 & \cellcolor{darkblue!25} 0.8523 \\
  & \cellcolor{darkblue} 0.8 & \cellcolor{darkblue!25} 0.8675 & \cellcolor{darkblue!25} 0.8494 & \cellcolor{darkblue!25} 0.8668 & \cellcolor{darkblue!25} 0.8677 & \cellcolor{darkblue!25} 0.8678 & \cellcolor{darkblue!25} 0.8662 & \cellcolor{darkblue!25} 0.8672 & \cellcolor{darkblue!25} 0.8661 \\
  & \cellcolor{darkblue} 1.0 & \cellcolor{darkblue!25} 0.8282 & \cellcolor{darkblue!25} 0.8293 & \cellcolor{darkblue!25} 0.8270 & \cellcolor{darkblue!25} 0.8272 & \cellcolor{darkblue!25} 0.8281 & \cellcolor{darkblue!25} 0.8263 & \cellcolor{darkblue!25} 0.8284 & \cellcolor{darkblue!25} 0.8278 \\
\bottomrule
\end{tabular}%
}
\end{table*}

\subsection{Generalization Analysis}
\hspace*{2em} To further evaluate the generalization capability of the proposed model across domains, experiments were conducted by training the model separately on the BraTS2018 and MU-Glioma-Post datasets and testing it on the unseen Pretreat-MetsToBrain-Masks dataset, as summarized in Table~\ref{table:cross_2018_in_pretreate} and Table~\ref{table:cross_mu_in_pretreate}. Taking the model trained on BraTS2018 as an example, the results on the Pretreat-MetsToBrain-Masks dataset show that AUROC values across different modality combinations range from 0.8580 to 0.8635, AP values range from 0.8842 to 0.8909, F1 scores range from 0.7730 to 0.7816, and AUPRO values range from 0.8332 to 0.8381. As shown in Fig.~\ref{fig:anomaly_score_distributions in different dataset} and Fig.~\ref{fig:anomaly_score_distributions in different combanation}, although BraTS2018 and Pretreat-MetsToBrain-Masks differ significantly in data distribution and acquisition conditions, the model maintains stable performance across all evaluation metrics, demonstrating remarkable cross-domain generalization. Similarly, when the model is trained on MU-Glioma-Post and tested on Pretreat-MetsToBrain-Masks, consistent results further confirm the robustness of the proposed framework. This special property establishes a highly solid and reliable foundation for effectively deploying the model in multi-source, heterogeneous medical imaging scenarios.

\begin{table*}[H]
\centering
\caption{\normalfont\normalsize Cross-domain validation results of the BraTS2018-trained model on the Pretreat-MetsToBrain-Masks dataset.}
\vspace{0.5em}
\label{table:cross_2018_in_pretreate}
\resizebox{\textwidth}{!}{%
\begin{tabular}{@{}l *{8}{c}@{}}
\toprule
\textbf{Metrics} &
\circletextgray{F}\circletext{T1}\circletext{T2} & 
\circletext{F}\circletextgray{T1}\circletext{T2} & 
\circletext{F}\circletext{T1}\circletextgray{T2} & 
\circletextgray{F}\circletextgray{T1}\circletext{T2} & 
\circletextgray{F}\circletext{T1}\circletextgray{T2} & 
\circletext{F}\circletextgray{T1}\circletextgray{T2} & 
\circletextgray{F}\circletextgray{T1}\circletextgray{T2} & 
\textbf{Avg} \\
\midrule
\multirow{1}{*}{\textbf{AUROC}} 
  & \cellcolor{lightyellow!25} 0.8611/0.9839
  & \cellcolor{lightyellow!25} 0.8584/0.9835
  & \cellcolor{lightyellow!25} 0.8609/0.9837
  & \cellcolor{lightyellow!25} 0.8620/0.9841
  & \cellcolor{lightyellow!25} 0.8599/0.9836
  & \cellcolor{lightyellow!25} 0.8580/0.9837
  & \cellcolor{lightyellow!25} 0.8635/0.9845
  & \cellcolor{lightyellow!25} 0.8606/0.9838 \\ 

\multirow{1}{*}{\textbf{AP}}
  & \cellcolor{lightred!25} 0.8881/0.6352
  & \cellcolor{lightred!25} 0.8844/0.6297
  & \cellcolor{lightred!25} 0.8872/0.6328
  & \cellcolor{lightred!25} 0.8889/0.6372
  & \cellcolor{lightred!25} 0.8865/0.6307
  & \cellcolor{lightred!25} 0.8842/0.6331
  & \cellcolor{lightred!25} 0.8909/0.6411
  & \cellcolor{lightred!25} 0.8869/0.6343 \\ 

\multirow{1}{*}{\textbf{F1}}
  & \cellcolor{lightpurple!25} 0.7779/0.6173
  & \cellcolor{lightpurple!25} 0.7745/0.6123
  & \cellcolor{lightpurple!25} 0.7776/0.6155
  & \cellcolor{lightpurple!25} 0.7787/0.6189
  & \cellcolor{lightpurple!25} 0.7774/0.6140
  & \cellcolor{lightpurple!25} 0.7730/0.6146
  & \cellcolor{lightpurple!25} 0.7816/0.6223
  & \cellcolor{lightpurple!25} 0.7770/0.6164 \\ 

\multirow{1}{*}{\textbf{AUPRO}}
  & \cellcolor{darkblue!25} 0.8355
  & \cellcolor{darkblue!25} 0.8328
  & \cellcolor{darkblue!25} 0.8349
  & \cellcolor{darkblue!25} 0.8368
  & \cellcolor{darkblue!25} 0.8348
  & \cellcolor{darkblue!25} 0.8332
  & \cellcolor{darkblue!25} 0.8381
  & \cellcolor{darkblue!25} 0.8350 \\ 
\bottomrule
\end{tabular}%
}
\end{table*}

\subsection{Zero-shot anomaly-detection}
\hspace*{2em}Finally, the model was evaluated in a zero-shot anomaly-detection scenario, as summarized in Table~\ref{table:zero_shot}. Although the proposed framework is not explicitly designed for zero-shot tasks, it achieves competitive performance on the previously unseen single-modality T1ce input scans, with image-level AP of {0.8688} and pixel-level AP of {0.9666}. These results demonstrate a certain degree of strong task transferability and robustness, and indicate that the proposed feature alignment and modality completion mechanisms possess inherent extensibility, providing a feasible direction for future applications in zero-shot or few-shot AD.

\section{Conclusion}
\hspace*{2em} In this study, we introduced a unified any-modality AD framework that addresses two fundamental limitations of existing medical AD methods: dependence on fixed modality configurations and poor generalization under real-world modality absence. By integrating dual-pathway DINOv2 feature extraction, a feature distribution alignment strategy, and INP-guided reconstruction, our model learns modality-invariant normal patterns and achieves robust anomaly detection across arbitrary MRI modality combinations. The proposed indirect feature-completion training further enables the model to compensate for missing information without explicit imputation or retraining.

\hspace*{2em} Comprehensive experiments on three heterogeneous glioma and metastasis datasets demonstrate that the framework not only surpasses state-of-the-art AD methods but also maintains stable performance across seven diverse modality settings, highlighting its strong generalizability and cross-domain robustness. Beyond methodological advances, this study establishes a scalable paradigm for multimodal medical image analysis under imperfect acquisition conditions, providing a practical and clinically aligned solution for real-world deployment. Future work will explore extending this paradigm to 3D volumetric AD, cross-institution domain shifts, and broader clinical applications where multimodal incompleteness is inherent.

\begin{table*}[h]
\centering
\caption{\normalfont\normalsize Cross-domain validation results of the MU-Glioma-Post-trained model on the Pretreat-MetsToBrain-Masks dataset.}
\vspace{0.5em}
\label{table:cross_mu_in_pretreate}
\resizebox{\textwidth}{!}{%
\begin{tabular}{@{}l *{8}{c}@{}}
\toprule
\textbf{Metrics} &
\circletextgray{F}\circletext{T1}\circletext{T2} & 
\circletext{F}\circletextgray{T1}\circletext{T2} & 
\circletext{F}\circletext{T1}\circletextgray{T2} & 
\circletextgray{F}\circletextgray{T1}\circletext{T2} & 
\circletextgray{F}\circletext{T1}\circletextgray{T2} & 
\circletext{F}\circletextgray{T1}\circletextgray{T2} & 
\circletextgray{F}\circletextgray{T1}\circletextgray{T2} & 
\textbf{Avg} \\
\midrule
\multirow{1}{*}{\textbf{AUROC}} 
  & \cellcolor{lightyellow!25} 0.8642/0.9844
  & \cellcolor{lightyellow!25} 0.8616/0.9842
  & \cellcolor{lightyellow!25} 0.8634/0.9844
  & \cellcolor{lightyellow!25} 0.8654/0.9847
  & \cellcolor{lightyellow!25} 0.8640/0.9844
  & \cellcolor{lightyellow!25} 0.8604/0.9845
  & \cellcolor{lightyellow!25} 0.8675/0.9851
  & \cellcolor{lightyellow!25} 0.8637/0.9845 \\ 
\multirow{1}{*}{\textbf{AP}}
  & \cellcolor{lightred!25} 0.8885/0.6251
  & \cellcolor{lightred!25} 0.8850/0.6218
  & \cellcolor{lightred!25} 0.8875/0.6231
  & \cellcolor{lightred!25} 0.8897/0.6298
  & \cellcolor{lightred!25} 0.8885/0.6237
  & \cellcolor{lightred!25} 0.8827/0.6219
  & \cellcolor{lightred!25} 0.8921/0.6340
  & \cellcolor{lightred!25} 0.8877/0.6241 \\ 

\multirow{1}{*}{\textbf{F1}}
  & \cellcolor{lightpurple!25} 0.7882/0.6137
  & \cellcolor{lightpurple!25} 0.7831/0.6110
  & \cellcolor{lightpurple!25} 0.7869/0.6124
  & \cellcolor{lightpurple!25} 0.7884/0.6170
  & \cellcolor{lightpurple!25} 0.7875/0.6128
  & \cellcolor{lightpurple!25} 0.7844/0.6121
  & \cellcolor{lightpurple!25} 0.7916/0.6201
  & \cellcolor{lightpurple!25} 0.7869/0.6140 \\ 

\multirow{1}{*}{\textbf{AUPRO}}
  & \cellcolor{darkblue!25} 0.8348
  & \cellcolor{darkblue!25} 0.8333
  & \cellcolor{darkblue!25} 0.8347
  & \cellcolor{darkblue!25} 0.8357
  & \cellcolor{darkblue!25} 0.8349
  & \cellcolor{darkblue!25} 0.8338
  & \cellcolor{darkblue!25} 0.8365
  & \cellcolor{darkblue!25} 0.8348 \\ 
\bottomrule
\end{tabular}%
}
\end{table*}

\begin{table}[H]
\centering
\caption{\normalfont\normalsize Zero-shot anomaly-detection results.}
\vspace{0.5em}
\label{table:zero_shot}
\resizebox{\columnwidth}{!}{%
\begin{tabular}{@{}l *{4}{c}@{}}
\toprule
 & BraTS2018 & MU & Brats\_in\_Pretreat & MU\_in\_Pretreat \\
\midrule
\multirow{1}{*}{\textbf{AUROC}}
& \cellcolor{lightyellow!25} 0.8618/0.9666
  & \cellcolor{lightyellow!25} 0.7700/0.9666
  & \cellcolor{lightyellow!25} 0.7122/0.9403
  & \cellcolor{lightyellow!25} 0.7315/0.9591 \\
\multirow{1}{*}{\textbf{AP}}
  & \cellcolor{lightred!25} 0.8942/0.5559
  & \cellcolor{lightred!25} 0.8226/0.4731
  & \cellcolor{lightred!25} 0.7351/0.3059
  & \cellcolor{lightred!25} 0.7648/0.3523 \\
\multirow{1}{*}{\textbf{F1}}
  & \cellcolor{lightpurple!25} 0.7830/0.5286
  & \cellcolor{lightpurple!25} 0.7068/0.4537
  & \cellcolor{lightpurple!25} 0.6869/0.3390
  & \cellcolor{lightpurple!25} 0.6922/0.3647 \\
\multirow{1}{*}{\textbf{AUPRO}}
  & \cellcolor{darkblue!25} 0.7707
  & \cellcolor{darkblue!25} 0.7910
  & \cellcolor{darkblue!25} 0.7453
  & \cellcolor{darkblue!25} 0.7664 \\
\bottomrule
\end{tabular}%
}
\end{table}

\section*{CRediT authorship contribution statement}

\hspace*{2em} \textbf{Changwei Wu}: Methodology, Visualization, Writing - original draft.  
\textbf{Yifei Chen}: Conceptualization, Methodology, Writing - original draft.  
\textbf{Yuxin Du}: Validation, Visualization.  
\textbf{Mingxuan Liu}: Validation, Visualization.  
\textbf{Jinying Zong}: Validation.  
\textbf{Jie Dong}: Visualization.   
\textbf{Beining Wu}: Visualization.  
\textbf{Feiwei Qin}: Supervision, Writing - review \& editing.  
\textbf{Yunkang Cao}: Writing - review \& editing.  
\textbf{Qiyuan Tian}: Project administration, Writing - review \& editing.

\section*{Declaration of competing interest}
\hspace*{2em} The authors declare that they have no known competing financial interests or personal relationships that could have appeared to influence the work reported in this paper.

\section*{Acknowledgments}
\hspace*{2em} This work was supported by the National Natural Science Foundation of China (No. 82302166), Tsinghua University Startup Fund, Fundamental Research Funds for the Provincial Universities of Zhejiang (No. GK259909299001-006), Anhui Provincial Joint Construction Key Laboratory of Intelligent Education Equipment and Technology (No. IEET202401), and the State Key Lab of CAD\&CG, Zhejiang University (A2510).

\bibliographystyle{elsarticle-num} 

\end{document}